%% file: main.tex
\documentclass[acmsmall, nonacm]{acmart}

\AtBeginDocument{%
  \providecommand\BibTeX{{%
    \normalfont B\kern-0.5em{\scshape i\kern-0.25em b}\kern-0.8em\TeX}}}

\usepackage{enumitem}
\usepackage{hyperref}
\usepackage{color,soul}

\include{bing-commands}

\usepackage{stmaryrd}
\usepackage[skip = 0.15pt]{subcaption} %
\usepackage{multirow}
\usepackage{wrapfig}
\usepackage{array}
\usepackage{algorithm}
\usepackage{algorithmic}
\usepackage{amsmath} % If not already loaded
\usepackage{amsthm}  % If not already loaded

\renewcommand\footnotetextcopyrightpermission[1]{} % Remove copyright notice
\begin{document}

\setlength{\abovedisplayskip}{0.96pt}
\setlength{\belowdisplayskip}{0.96pt}

\title{Cross-platform Prediction of  Depression Treatment Outcome Using Location Sensory Data on Smartphones}
%%%==================================================

\author{Soumyashree Sahoo, Chinmaey Shende, Md. Zakir Hossain}
 \email{firstname.lastname@uconn.edu}
\affiliation{%
	\institution{University of Connecticut}
	\streetaddress{School of Computing}	 
	\city{Storrs}
	\state{CT}
%	\postcode{06269}
	\country{USA}
}
\author{Parit Patel}
 \email{papatel@uchc.edu}
\affiliation{%
	\institution{University of Connecticut Health}
	\streetaddress{Department of Psychiatry}	 
	\city{Farmington}
	\state{CT}
%	\postcode{06269}
	\country{USA}
}
\author{Yushuo Niu, Xinyu Wang}
 \email{firstname.lastname@uconn.edu}
\affiliation{%
	\institution{University of Connecticut}
	\streetaddress{School of Computing}	 
	\city{Storrs}
	\state{CT}
%	\postcode{06269}
	\country{USA}
}
\author{Shweta Ware}
 \email{sware@richmond.edu}
\affiliation{%
	\institution{University of Richmond}
	\streetaddress{Department of Computer Science}	 
	\city{Richmond}
	\state{VA}
%	\postcode{06269}
	\country{USA}
}
\author{Jinbo Bi}
 \email{jinbo.bi@uconn.edu}
\affiliation{%
	\institution{University of Connecticut}
	\streetaddress{School of Computing}	 
	\city{Storrs}
	\state{CT}
%	\postcode{06269}
	\country{USA}
}
\author{Jayesh Kamath}
 \email{jkamath@uchc.edu}
\affiliation{%
	\institution{University of Connecticut Health}
	\streetaddress{Department of Psychiatry}	 
	\city{Farmington}
	\state{CT}
%	\postcode{06269}
	\country{USA}
}
\author{Alexander Russel, Dongjin Song, Qian Yang, Bing Wang}
 \email{acr@uconn.edu, dongjin.song@uconn.edu, qyang@uconn.edu, bing@uconn.edu}
\affiliation{%
	\institution{University of Connecticut}
	\streetaddress{School of Computing}	 
	\city{Storrs}
	\state{CT}
%	\postcode{06269}
	\country{USA}
}

%------------------------------------------------------------------------
\input{abstract}

% \begin{CCSXML}
% <ccs2012>
%    <concept>
%        <concept_id>10002951.10003227.10003245</concept_id>
%        <concept_desc>Information systems~Mobile information processing systems</concept_desc>
%        <concept_significance>300</concept_significance>
%        </concept>
%  </ccs2012>
% \end{CCSXML}

% \ccsdesc[300]{Information systems~Mobile information processing systems}

% \keywords{Depression, Mental Health, Machine Learning, Sensory Data 
% }

%%%%%%%% updated %%%%%%%%%%%%%%%
\input{abstract}
\maketitle

\input{intro}

\input{related}

\input{data}

\input{data-preprocessing}
\input{Feature-Extraction}

\input{DomainAdaptation-New}

\input{correlation-analysis}   % start here to update
\input{predict}

\input{discuss-SS}
\input{conclusion-SS}

%%%%%%%%%%%%%%%%%%%%%%%%%%%%%%%%%

\bibliographystyle{abbrv}
\bibliography{bib/ref-new}

\end{document}

%% file: bing-commands.tex
\usepackage{epsfig}
\usepackage{algorithm}
\newcommand{\remove}[1]{}

\newtheorem{lemma}{Lemma}
\newtheorem{corollary}{Corollary}
\newcommand{\ls}[1]
   {\dimen0=\fontdimen6\the\font
    \lineskip=#1\dimen0
    \advance\lineskip.5\fontdimen5\the\font
    \advance\lineskip-\dimen0
    \lineskiplimit=.9\lineskip
    \baselineskip=\lineskip
    \advance\baselineskip\dimen0
    \normallineskip\lineskip
    \normallineskiplimit\lineskiplimit
    \normalbaselineskip\baselineskip
    \ignorespaces
   }

%% file: abstract.tex
\begin{abstract}

\vspace{0.2cm}
\textbf{ABSTRACT}\\[0.2em]
Currently, depression treatment relies on closely monitoring patients' response to treatment and adjusting the treatment as needed. Using self-reported or physician-administrated questionnaires to monitor treatment response is, however, burdensome, costly and suffers from recall bias. In this paper, we explore using location sensory data collected passively on smartphones to predict treatment outcome.  To address heterogeneous data collection on Android and iOS phones, the two predominant smartphone platforms, we explore using domain adaptation techniques to map their data to a common feature space, and then use the data jointly to train machine learning models. Our results show that this domain adaptation approach can lead to significantly better prediction than that with no domain adaptation. In addition, our results show that using location features and baseline self-reported questionnaire score can lead to $F_1$ score up to 0.67, comparable to that obtained using periodic self-reported questionnaires, indicating that using location data is a promising direction for predicting depression treatment outcome.     
\end{abstract}

%% file: intro.tex
\section{Introduction} 

Depression is a highly prevalent and debilitating mental health disorder that can have significant impacts on both individuals and society as a whole~\cite{Cuijpers02:Excess,katon2002impact}. 
Improving the recognition and treatment of depression is essential for reducing its burden and improving overall public health~\cite{doi:10.1056/NEJMp2008017,07648d6cc15940829f39edccd66719ab}.
However, very few clinical characteristics, biomarkers, or genetic variations have been identified that can reliably predict differential effectiveness of specific depression treatments~\cite{Kemp08:improving,Simon10:personalized,Cohen18:treatment}. At a result, it remains difficult to find the perfect treatment for individual patients, and the best approach thus far is closely monitoring the treatment status, assessing depression symptoms over time, and adjust the treatment
as needed~\cite{Morris12:Measurement-Based,Fortney17:Measurement-Based}.

The current methods for assessing depression symptoms rely on self-reported or physician-administrated questionnaires, which have multiple
 limitations such as long intervals between assessments, recall bias, and social desirability bias~\cite{Zeev10:Retrospective,Stone02:Capturing}. It is crucial to have objective, accurate, and timely assessments to help physicians provide personalized treatment for patients with depression. Mobile devices such as smartphones and wearables can be used to collect sensory data passively, which can be used for long-term monitoring of behavioral manifestations of depression symptoms, without relying on burdensome questionnaires  
 (see \S\ref{sec:related}). %Passive sensory data collected by these devices can provide critical information that correlates with depression symptoms. 
 However, most existing studies focus on detecting depression onset or relapse; there is much less work on predicting improvement or lack of improvement of depression symptoms  using sensory data over time to guide depression treatment.

In this paper, we explore using location sensory data collected from smartphones to predict depression treatment outcome, i.e., whether a patient is improving or not after initiating  treatment. The premise of our study is that location data can be used to infer a rich set of behavioral features such as  regularity of movement patterns, variance of locations visited, and proportion of time spent at home, which have been shown to be correlated with depression symptoms~\cite{Canzian:2015:TDU:2750858.2805845,wang2014studentlife,saeb2015mobile,Farhan16:Multi-view,Farhan16:depression,Yue18:fusion,Darius18:Correlations}. One challenge we face is that location features derived from sensory data collected on Android and iOS phones, the two primary smartphone platforms, are not compatible, due to various system related differences and differences in data collection mechanisms  (see \S\ref{sec:data-collection}). Developing machine learning models using the data from each platform individually will lead to reduced sample size, diminishing the power of any analysis. The study in~\cite{Lu2018:MTL} addresses this issue using a multi-task learning framework, which is not suitable for our study since we address a single-task problem (i.e., predicting treatment status).
Instead, we adopt {\em domain adaptation}~\cite{Pan10:transfer-learning,Shai10:domain-adapt,Redko19:domain-adapt} to align the datasets and then use the datasets jointly to train prediction models. Using a dataset from a community sample of 66 participants, our study makes the following main contributions:

\begin{itemize}
    \item We explore a novel direction that uses domain adaptation to address heterogeneous data collection on different platforms. Specifically, we use the domain adaptation approach in~\cite{Sun15:CORAL} to transform the iPhone and Android features into the same feature space to facilitate later machine learning tasks. Specifically, we explore three forms of domain adaptation: {\em Android-transformed}, which %treats Android dataset as source, and iOS dataset as target, and 
    transforms Android dataset to be in the feature space of the iOS dataset, {\em iOS-transformed}, which transforms iOS dataset to be in the feature space of the Android dataset, and {\em dual-transformed}, which transforms Android and iOS datasets jointly  into a common feature space. Our evaluation shows that all three forms of domain adaptation are effective in aligning the distributions of the location features extracted from these two platforms.

    \item We use the transformed data from the two platforms to train a family of machine learning models, based on Support Vector Machine~\cite{CC01a:libSVM} and  XGBoost~\cite{chen2016xgboost}, to predict depression treatment outcome. Specifically, we investigate prediction in multiple scenarios, including using self-reported scores alone, using current location data alone, incorporating location baseline, and incorporating baseline self-reported scores. Our results show that using the combined transformed data leads to much better prediction accuracy than using the combined data without domain adaptation. In addition, the dual-transformed approach leads to better results than the other two forms of domain adaptation.
        
    \item Our results show that location sensory data can provide alternative means for predicting depression treatment outcome, without the help of burdensome periodic questionnaire responses.  
    Using location features that can be gathered easily and baseline self-reported questionnaire score that is routinely collected at the beginning of depression treatment, 
    the machine learning models can provide $F_1$ score up to 0.67, comparable to the best $F_1$ score (0.69) obtained using periodic self-reported scores.
\end{itemize}

The rest of the paper is organized as follows. We briefly review related work in Section~\ref{sec:related}. We then present data collection and pre-processing in Sections~\ref{sec:data-collection} and \ref{sec:data-preprocessing}, respectively. After that, we present feature extraction, cross-platform domain adaptation, and correlation analysis in Section~\ref{sec:corr}. 
We present machine learning based prediction in Section \ref{sec:class}. Discussion and limitation of this work are presented in Section~\ref{sec:discuss}. Finally, Section~\ref{sec:conclusions} concludes the paper.

%% file: related.tex
\section{Related Work} \label{sec:related}

\emph{\noindent{\bf Predict depression treatment outcome and severity changes.}} Our work is in the category of predicting depression treatment outcome, specifically, whether the depression symptom severity level has improved or not, after initiating a treatment. While there is extensive research on this topic, only recent studies have developed machine learning models~\cite{Lee18:predict,Chekroud21:promise,Rost22:treatment}, and most of them utilized baseline clinical data, instead of sensory data that can be continuously collected. A recent study~\cite{Dai22:treatment} used  baseline clinical characteristics along with the first 2-month sensory data collected by wearable devices to predict the efficacy of a new depression treatment for individual patients. It proposed a multi-task learning model that is trained on both intervention and control groups. Another study~\cite{Zou23:Sequence} uses sensory data collected by smartphones and wearables in the first 2-4 weeks of the treatment to predict the outcome for a later time (12th week).
The study  in~\cite{Shende2023:PredictingSymptom} uses sleep data collected from Fitbit to predict treatment improvement. Another recent study~\cite{Sahoo2024:DailyMood} uses daily mood and anxiety survey collected from smartphones to predict treatment improvement. 
 Our work differs from them in that we use location sensory data collected from smartphones, and address the incompatibility of the data on different platforms using domain adaptation. 

Several studies~\cite{Ben-Zeev15L:psychiatric,meyerhoff2021evaluation,Demasi16:change,Chikersal21:depression} used sensory data to predict depression severity level changes.  Unlike our study, these existing studies are not in clinical settings; they do not use clinician assessment as ground truth, and do not predict treatment outcome after patients initiates new treatments. Most of these studies use 
a variety of sensor modalities, instead of only location data as in this study. 
As our study, the study in~\cite{Canzian:2015:TDU:2750858.2805845} also only considered mobility data.
It used PHQ score~\cite{Spitzer99:PHQ-9} as the ground truth, and trained personalized and general machine learning models to predict whether the current PHQ score exceeds the average PHQ score added by one standard deviation. Our study differs from \cite{Canzian:2015:TDU:2750858.2805845} in that we use clinical assessment, not self-reported scores, as the ground truth, and handle location data from different platforms.

\emph{\noindent{\bf Predict depression using sensory data.}} A large number of 
recent studies have used sensing data (e.g., physical activity, location, sleep) collected on smartphones and/or wearables  for detecting depression or depressive mood~\cite{frost2013supporting,gruenerbl2014using,grunerbl2012towards,Canzian:2015:TDU:2750858.2805845,wang2014studentlife,saeb2015mobile,Ben-Zeev15L:psychiatric,Mehrotra16:Towards,Zhou15:tackling,Wang16:schizophrenia,Palmius16Detecting,Farhan16:Multi-view,Farhan16:depression,Suhara16:DeepMood,Chow17:mobilesensing,Yue18:fusion,Lu2018:MTL,Wang18:dynamics,Zhang21:Relationship,Chikersal21:depression}.
These studies extract behavioral features from the sensing data, show that they are correlated with depression symptoms, and develop machine learning models or statistical techniques to predict depression~\cite{Mohr17:sensing}. 
Our work differs from them in that we focus on predicting depression treatment outcome, instead of the the onset of relapse of depression.

\emph{\noindent{\bf Domain adaptation.}} Our study leverages existing domain adaptation techniques to align the distributions of Android and iPhone location features. 
The main purpose of domain adaptation is adapting a machine learning model that is trained on data from one domain to perform well on data from a different, but related domain~\cite{Pan10:transfer-learning,Shai10:domain-adapt,Redko19:domain-adapt}. The goal is to overcome the problem of distributional shift, where the distribution of the input data in the target domain may be different from that in the source domain. In this paper, we use the technique in~\cite{Sun15:CORAL} to transform data from two smartphone platforms (Android and iOS phones)  into one common feature space, and then use them jointly to train machine learning models. Specifically, we explore three forms of domain adaptation (see \S\ref{sec:domain-adaptation}); the dual-transformed approach  extending the technique in~\cite{Sun15:CORAL}.

%% file: data.tex
\section{Data Collection}
\label{sec:data-collection}
We collected sensory data and self-reported questionnaire scores on smartphones.
%, along with  clinical assessment. 
Each participant is assigned a random user ID for this study. The
data collected by the app is only associated with the random ID. It is encrypted before being stored on
the phone, and then sent to a secure server to protect user privacy. In the following, we first describe participant recruitment and then data collection.

\emph{\noindent{\bf Recruitment.}} 
The participants of this study were recruited from January 2020 to December 2023,  from several mental health clinics. Based on the enrollment criteria, all the participants were diagnosed with depression, at least 18 years old, English speaking, and starting a new pharmacological treatment for depression (i.e., starting a new medication  or increasing the dose of the current medication). Participants who had any co-morbid severe mental illness such as bipolar disorder, schizophrenia, or other psychotic disorders were excluded from the study. The study protocols and procedures were approved by the Institutional Review Board (IRB) of the University of Connecticut. 
All participants met with our study clinician for informed consent and initial screening before being enrolled in the study.

We recruited a total of 104 participants for this study.
The participants use either Android or iOS phones (iOS is the operating system of iPhones; we use iOS phone and iPhone interchangeably in this paper). Specifically, 31 used Android and 73 used iOS phones. Out of the 31 Android users, 3 withdrew during the first week of study, 3 had not responded to  monthly followup assessments. Out of 73 iOS users, 9 withdrew within a few days of study, 6 had not responded to monthly followup assessments. Summarizing the above, the data analysis below is for 25 Android and 58 iOS users. 

All participants use their own smartphones (either iOS or Android).
Overall, our study group has 25 Android users, and 58 iPhone users. Since almost all the participants used their own phone, we expect to collect data with a reasonably good quality, as people tend to carry and actively use their own phones.

\emph{\noindent{\bf Self-report Questionnaire.} }
We used Quick Inventory of Depressive Symptomatology (QIDS)~\cite{rush200316}, a widely used self-assessment questionnaire, for this study. QIDS 
measures 16 factors across 9 different criterion domains including  mood, concentration, self-criticism, suicidal ideation, interests,  energy/fatigue, sleep disturbance,
decrease or increase in appetite or weight, and psychomotor agitation or retardation. The total score of QIDS ranges from 0 to 27; higher scores indicate higher severity. The participants filled in QIDS at the beginning of the study, which were treated as their {\em baseline QIDS score}. Only those with baseline QIDS score $\ge 11$  were recruited into the study, since QIDS score of 11 is often used as a cutoff value that indicates moderate depression. Once enrolled, participants filled in QIDS every 7 days on their phones. A notification was sent to their phones on the due date.

\emph{\noindent{\bf Clinical Assessment.}} Our study clinician screened the participants at the enrollment time and end of each month to determine the corresponding Clinical Global Impressions (CGI)~\cite{Guy76:CGI} score.  CGI comprises two companion one-item measures. One is CGI-S that evaluates the severity of psychopathology from 1 (normal) to 7 (amongst the most extremely ill patients). The other is CGI-I that evaluates the improvement/change of the symptoms relative to the baseline (i.e., the initiation of the new or increased medication in our context) on a similar seven-point scale, from 1 (very much improved) to 7 (very much worse). In the rest of the paper, we use  CGI-I score as the ground truth for patient treatment improvement status. CGI-I value 1 (very much improved) or 2 (much improved) is considered as {\em improved}, while the other values (i.e., 3-7, corresponding to minimally improved to very much worse) is considered as {\em not improved}.

\input{sensory-data}

%% file: sensory-data.tex
{\bf Sensory Data.} We collected location sensory data on both Android and iOS platforms. 
While Android allows periodic data collection in the background, iOS has much stricter rules about collecting data in the background. Several mobile sensing frameworks~\cite{katevas2016sensingkit,Xiong2016SensusAC,Yuuki2020:Aware,Farhan16:depression} can collect location data for iOS. Among them, AWARE-iOS~\cite{Yuuki2020:Aware} %developed  that builds on top of the \bing{AWARE~\cite{}} sensing platform for iOS. While AWARE-iOS
allows periodic sensing data collection for iOS. It, however, relies on a push notification service from a remote server to sustain the data collection on a phone when there is no frequent user activities with the app. In this study, we chose to use LifeRhythm app~\cite{Farhan16:depression}, since our study requires minimum interaction with users, and some users do not have cellular data services to maintain consistent connection with a remote server (for push notification).

Specifically, we used the app to collect two types of location data, GPS and WiFi association data; the latter is relevant since, if a phone is associated with an AP, then we can use the location of the AP to approximate the location of the user.
After that, we used the approach in \cite{Yue18:fusion} to fuse GPS and WiFi data
to  obtain more complete location information. We next briefly describe data collection and then the location fusion approach. 

{\em GPS data collection.} 
LifeRhythm  app~\cite{Farhan16:depression} collects GPS data using different mechanisms on Android and iOS platforms due to the different restrictions of their operating systems. 
On Android phones, GPS location was collected {\em periodically} every 10 minutes.  
On iOS phones, locations were collected using an event based mechanism, since iOS does not provide APIs to schedule periodic data collection. Specifically, the app subscribes to the location services provided by the operating system and 
obtained location updates after a user has traveled a certain distance (which is set to 50 meters to 1 mile based on user activity)~\cite{Farhan16:depression}. When such an event occurs, the app will sense and record the event. The desired accuracy is switched between 10 and 100 meters depending on user’s activity to achieve accurate location collection,   while minimizing the impact on battery life (higher accuracy leads to more energy consumption)~\cite{Farhan16:depression}. Each location sample contains  longitude, latitude, user ID, and error (in meters). Following~\cite{Farhan16:depression}, we removed the samples that had errors larger than 165 meters to retain most of the samples while eliminating the samples with large errors.

{\em WiFi association data.} We also used LifeRhythm app to collect WiFi association and dissociation events. The entry for each event includes a timestamp and MAC address of the access point (AP), which serves as the unique identifier of the AP.

\input{fusion}

%% file: fusion.tex
{\em Fusing GPS and WiFi data.} The goal of data fusion is combining GPS and WiFi association data to output  a sequence of locations (in longitude and latitude coordinates) for each user on each day.  As described in \cite{Yue18:fusion}, it contains two steps: (i) estimate the longitude and latitude coordinates for the APs, and (ii) fuse GPS and WiFi location samples together.

To estimate the longitude and latitude coordinates for an AP $a$, we consider all the association events for $a$ from a user and the GPS samples that were collected from the same user during a time interval  5 minutes before and after each association event, and used the mean of the GPS coordinates as the location of $a$. Correspondingly, we obtained the geographic locations of 6054 and 670 APs for iOS and Android users, respectively.

\begin{figure}
     \centering
     \begin{subfigure}[b]{0.32\textwidth}
        %  \centering
         \includegraphics[width=\textwidth]{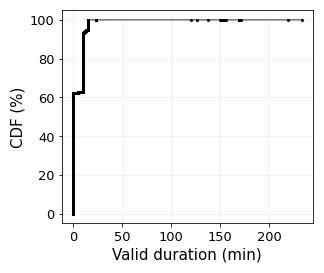}
         \caption{Android}
         \label{fig:AndroidTW}
     \end{subfigure}
     \begin{subfigure}[b]{0.32\textwidth}
        %  \centering
         \includegraphics[width=\textwidth]{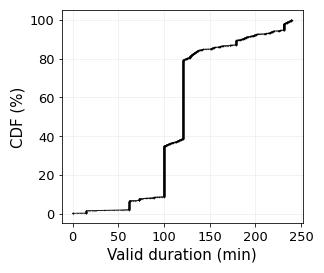}
         \caption{iOS}
         \label{fig:iOSTW}
     \end{subfigure}
    \caption{Time duration for which a location measurement is valid when merging GPS and WiFi location data (considering the data collected 6am - 10pm each day). %\bing{Soumya, can you make the text font size larger? x-axis: Valid duration (min). \soumya{updated}}
    } 
    \label{fig:locationTW}
\end{figure}

In the second step,  we consider a sequence of time points $\{t_i\}$, where each time point  $t_i$ has a location sample (obtained from GPS or WiFi). We determine the duration for which the location at $t_i$ is valid
following the approach in \cite{Yue18:fusion}. Specifically, we consider two thresholds, $T_G$ and $T_W$, for GPS data and WiFi respectively. If the location for $t_i$ is obtained using GPS, then the duration for which the user is assumed to be at this location is $[t_i, \min(t_i+T_G, t_{i+1})]$. Similarly, if the location for $t_i$ is obtained using WiFi, then the duration for which the user is assumed to be at this location is $[t_i, \min(t_i+T_W, t_{i+1})]$.   
For Android, $T_G$ is set to 15 minutes, and $T_W$ is set to  4 and 6 hours for weekdays and weekends respectively for 6am to 10pm; and set to 8 hours otherwise.
For iOS, $T_G$ and $T_W$ are both set as $T_W$ for Android. Fig.~\ref{fig:locationTW}a and b plot the distribution of 
the duration for which a location measurement is valid for Android and iOS, respectively. They are for the data collected during 6am - 10pm. As expected, for Android, these intervals tend to be within 15 minutes, since the GPS location for Android is collected periodically every 10 minutes. For iOS, these intervals are more widely spread due to the event-based data collection. On the other hand, {$81\%$} of the intervals are within {130} minutes.  After that, we use upsampling to obtain location data at 1-minute intervals, which will be used for feature extraction (see \S\ref{sec:corr}). More details of the data fusion methodology are found in \cite{Yue18:fusion}.

%% file: data-preprocessing.tex
\section{Data Pre-processing} \label{sec:data-preprocessing}
The analysis in the rest of the paper is on {\em QIDS intervals}. Each QIDS interval ends with the day when a participant fills in  QIDS questionnaire and the previous 7 days, since QIDS asks about the behaviors in the past 7 days. The location data associated with a QIDS interval is obtained through smartphone sensing as described earlier.

\begin{figure}
     \centering
     \begin{subfigure}[b]{0.24\textwidth}
         \includegraphics[width=\textwidth]{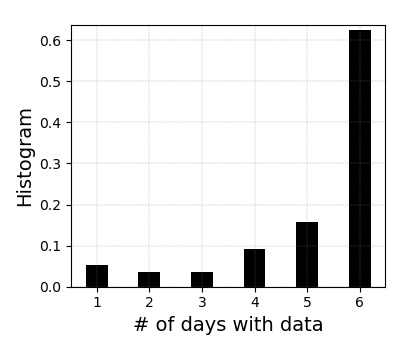}
         \caption{Android QIDS length}
         \label{fig:Android QIDS length}
     \end{subfigure}
     \begin{subfigure}[b]{0.24\textwidth}
         \includegraphics[width=\textwidth]{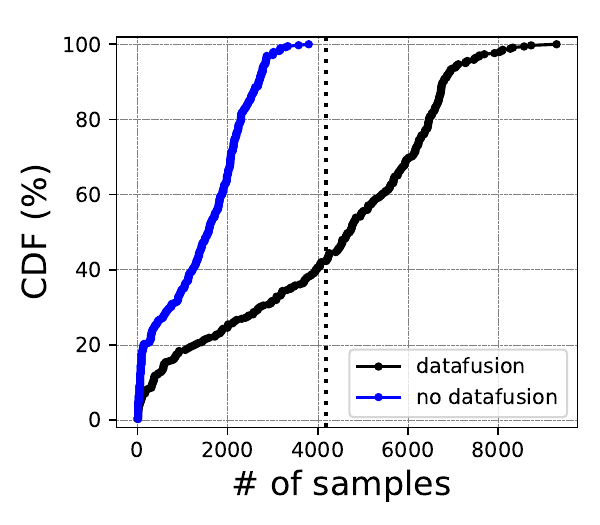}
         \caption{Android samples}
         \label{fig:Android samples}
     \end{subfigure}
     \begin{subfigure}[b]{0.24\textwidth}
        %  \centering
         \includegraphics[width=\textwidth]{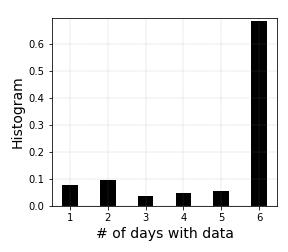}
         \caption{iOS QIDS length}
         \label{fig:iOS QIDS length}
     \end{subfigure}
     \begin{subfigure}[b]{0.24\textwidth}
        %  \centering
         % \includegraphics[width=\textwidth]{figures/preprocessing_location/samples_IOS-upsample.pdf}
         \includegraphics[width=\textwidth]{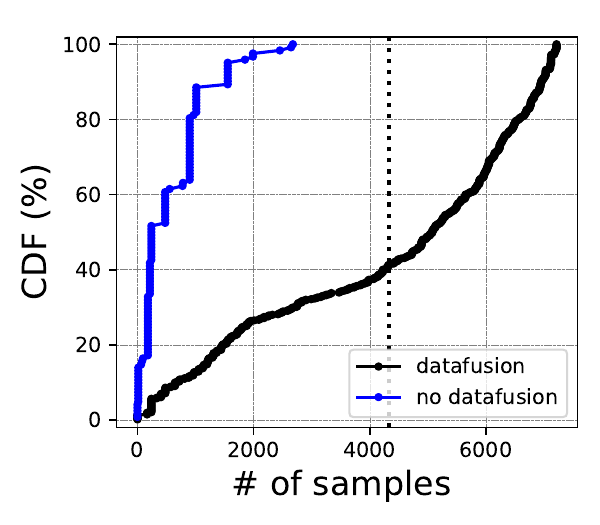}
         \caption{iOS samples}
         \label{fig:iOS samples}
     \end{subfigure}
    \caption{(a)-(b) show the   
    distributions of the number of days with location data and the number of location samples in a QIDS interval for the Android dataset. (c)-(d) show the corresponding distributions for the iOS dataset. }
\end{figure}

\emph{\noindent{\bf Missing Data.}} Despite the data fusion procedure that combines GPS and WiFi data, we still observed a significant amount of missing data, and sometimes no location data was collected at all in a day.  
Fig.~\ref{fig:Android QIDS length}  plots the histogram  of the number of days with location data in a QIDS interval for the Android dataset.  We see that 78\%   of the QIDS intervals have at least 5 days of data, 63\%  of the QIDS intervals have 6 days of data, while no QIDS interval has 7 days of data (the maximum number of days with data). Fig.~\ref{fig:Android samples} plots the cumulative distribution function (CDF) of the number of samples in the QIDS intervals for the Android dataset. Given the earlier observation that there are at most 6 days with data in a QIDS interval in the Android dataset, the maximum number of samples in a QIDS interval is $60 \times 24\times 6=8,640$,  since the location data is at 1-minute intervals. Most of the samples are below this value. On the other hand, we see that 76\% 
of the QIDS intervals have at least 2000 samples, significantly higher compared to that without merging GPS data with WiFi, which have only 34\%  of the QIDS intervals with at least 2000 samples (blue curve  in Fig.~\ref{fig:Android samples}). Considering that we need to have reasonable number of QIDS intervals and yet each QIDS interval needs to have a reasonable number of samples for feature extraction and data analysis, we excluded all QIDS intervals that have less than 5 days of data or have less than 2000 of samples. After applying the above criteria, out of the 25 Android users, 4 users were excluded since they did not have any QIDS interval that satisfies the above criteria.

Fig.~\ref{fig:iOS QIDS length} and  Fig.~\ref{fig:iOS samples} plot the histogram of the number of days with data and CDF of the number of samples in a QIDS interval for the iOS dataset, respectively. We similarly observe a significant amount of missing data: although 73\%  of the QIDS intervals have at least 5 days of data, no QIDS interval has 7 days of data, and 73.5\% of the QIDS intervals have more than 2000 samples. This is significantly higher compared to that without merging GPS and WiFi data, which only has 11\% (blue curve in Fig.~\ref{fig:iOS samples}c)  of the QIDS intervals with at least 2000 samples. 
When using the same filtering criteria as that used for the Android dataset, 13 out of the 58 iOS users were not included in further data analysis.

Summarizing the above, the rest of the paper considers the data from these 66 users (21 Android and 45 iOS users). Of them, 82.81\% were female and 17.19\% were male. In terms of ethnicity, they were 65.62\% white, 10.93\% Asian, 6.25\% African American, and 17.2\% had more than one race. Each participant was in the study for up to 12 weeks. 
For Android users, the days of participation varies from 32 to 84 days  with a mean of 72 days;  for iOS users, it varies from 32 to 84 days with a mean of 76 days.

\begin{figure}
     \centering
     \begin{subfigure}[b]{0.32\textwidth}
        %  \centering
         \includegraphics[width=\textwidth]{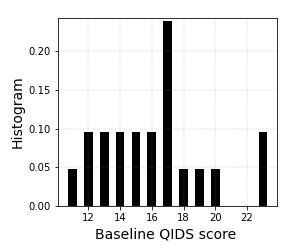}
         \caption{Android}
         \label{fig:Baseline QIDS Android}
     \end{subfigure}
    %  \hfill
     \begin{subfigure}[b]{0.32\textwidth}
        %  \centering
         \includegraphics[width=\textwidth]{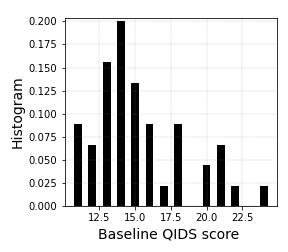}
         \caption{iOS}
         \label{fig:Baseline QIDS iOS}
     \end{subfigure}
    \caption{Baseline QIDS score for Android and iOS users. }
    %The vertical line in each plot represents the mean value. }
    \label{fig:baseline-QIDS}
\end{figure}

\begin{figure}
     \centering
     \begin{subfigure}[b]{0.32\textwidth}
        %  \centering
         \includegraphics[width=\textwidth]{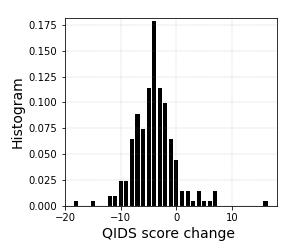}
         \caption{Android}
         \label{fig:QIDS variation Android}
     \end{subfigure}
    %  \hfill
     \begin{subfigure}[b]{0.32\textwidth}
        %  \centering
         \includegraphics[width=\textwidth]{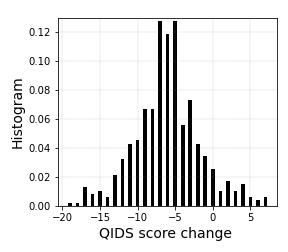}
         \caption{iOS}
         \label{fig:QIDS variation iOS}
     \end{subfigure}
    \caption{Histogram of  QIDS score changes for Android and iOS users.}
    \label{fig:QIDS-changes}
\end{figure}

\emph{\noindent{\bf QIDS Scores.}} Fig.~\ref{fig:baseline-QIDS} plots the baseline QIDS score (i.e., the QIDS score at the enrollment) for the Android and iOS users. We see that for Android users, the baseline QIDS score varies from 11 to 23 with a mean of 16.0. For iOS users, the variation is similar, from 11 to 24 with a mean of 15.3.
Fig.~\ref{fig:QIDS-changes} plots the distributions of QIDS score changes (i.e., a collected QIDS score subtracted by the baseline QIDS score) for Android and iOS users. We see that most of the score changes are negative, indicating less severe depression symptoms after the enrollment. The average changes for Android and iOS users are -2.8 and -6.1,  respectively. Particularly, for Android users, 30\% of the QIDS scores are more than 5 points below the baseline value; for iOS users, the corresponding value is 56.7\%. A small fraction of the score changes is positive.

\emph{\noindent{\bf Improvement Status.}} As mentioned earlier, we use CGI-I score as the ground truth to classify the improvement status for each QIDS interval. 
Specifically, suppose a CGI is obtained for a participant on day $t$, and the previous CGI is obtained on day $t'$, or $t'$ is the enrollment day. 
If the CGI on day $t$ indicates improved status, then we refer to the time period between day $t'$ and $t$ as improved. We define not-improved periods similarly.
For one participant, the improvement status may be stable over the entire duration of the study (i.e., remain improved or not-improved), or change over time. For the 21 Android participants, only 5 participant had change in improvement status from not-improved to improved, where as 2 participants have change in other direction. For the 45 iOS participants, 9 participants had one change in improvement status (9 had the change from not-improved to improved, and 2 had the change in the opposite direction).
Following the above procedure, for the Android dataset, 39 QIDS intervals (from 9 participants) are marked as improved, while 163 QIDS intervals (from 19 participants) are marked as not-improved. For the iOS dataset, 194 QIDS intervals (from 24 participants) are marked as improved, while 268 QIDS intervals (from 32 participants) are marked as not-improved. 
In summary, the Android dataset contains 202 samples from 21 users, while the iOS dataset contains 462 samples from 45 users. 

%% file: Feature-Extraction.tex
\section{Feature Extraction, Domain Adaptation, and Correlation Analysis} \label{sec:corr}

In this section, we present feature extraction to extract features from the location data, domain adaption to align the Android and iOS feature space. At the end, we correlate the location features with QIDs scores.

\subsection{Location Features}
We extract 8 location features from the location data for each QIDS interval. These features are similar to those in~\cite{saeb2015mobile,Saeb16:student-life-data,Farhan16:depression,Yue18:fusion}. Specifically, the first four features are directly based on location data, while the last four features are obtained based on locations clusters. Specifically, we use DBSCAN~\cite{ester1996density}, a density based clustering algorithm to cluster the stationary points. DBSCAN requires two parameters, epsilon (the distance between points) and the minimum number of points that can form a cluster (i.e., the minimum cluster size). Following the approach in~\cite{Yue18:fusion}, we set epsilon to 20 meters and the minimum number of points to 2.5 hours of stay (i.e., 160 points as two consecutive locations are one minute apart).

\emph{\noindent{\bf Location variance:} }
This feature measures the variability in a participant's location. It is calculated as $\log(\sigma^2_{\text{long}} + \sigma^2_{\text{lat}})$, where $\sigma^2_{\text{long}}$ and $\sigma^2_{\text{lat}}$ represent the  variance of the longitude and latitude of the location coordinates, respectively~\cite{saeb2015mobile}.

\emph{\noindent{\bf Time spent in moving:} }
This feature represents the percentage of time that a participant is moving. We differentiate moving and stationary samples using the approach in~\cite{saeb2015mobile}. Specifically, we estimate the moving speed at a sensed location. If the speed is larger than 1 km/h, then we classify it as moving; otherwise, we classify it as stationary.

\emph{\noindent{\bf Total distance:}}
Given the longitude and latitude of two consecutive location samples for a participant, we use Harversine formula  to calculate the distance traveled in kilometers between these two samples. The total distance traveled during a time period
is the total distance normalized by the time period~\cite{saeb2015mobile}.

\emph{\noindent{\bf Average moving speed:}}
This feature represents the average moving speed, where movement and speed are identified in the same way as what is used for the total distance feature.

\emph{\noindent{\bf Number of unique locations:}}
It is the number of unique clusters from the DBSCAN algorithm, denoted as $N_{loc}$~\cite{Farhan16:depression}

\emph{\noindent{\bf Entropy:}}
It measures the variability of time that a participant
spends at different locations~\cite{saeb2015mobile}. Let $p_i$ denote the percentage of
time that a participant spends in location cluster $i$. The entropy
is calculated as $Entropy = -\sum_i \ (p_i \log{p_i})$.

\emph{\noindent{\bf Normalized entropy:}}
It is $Entropy/\log N_{loc}$, and hence is invariant to the number of clusters and depends solely on the distribution of the visited location clusters~\cite{saeb2015mobile}.

\emph{\noindent{\bf Time spent at home:}}
 We use the approach in~\cite{saeb2015mobile,Farhan16:depression}  to identify ``home'' for a participant as the location cluster that the participant is most frequently found between $[12,6]$am. After that, we calculate the percentage of time when a participant is at home.

%% file: DomainAdaptation-New.tex
\subsection{Cross-platform Domain Adaptation} \label{sec:domain-adaptation} 

The Android and iOS datasets that we collected are not compatible due to different data collection mechanisms (see \S\ref{sec:data-collection}). As a  result, the location features extracted from these two datasets have significantly different distributions. 
We next explore using domain adaptation to transform the Android and iOS feature space to be compatible, which can then be combined to form larger training sets to train machine learning models (see \S\ref{sec:class}). We also present the prediction results of directly combining the Android and iOS datasets without any domain adaptation in \S\ref{sec:class}, and show that its performance can be significantly worse than those with domain adaptation.  

Domain adaptation~\cite{Pan10:transfer-learning,Shai10:domain-adapt,Redko19:domain-adapt}, which is broadly in the area of transfer learning, can be used to align two domains that have different distributions. While many domain adaptation approaches have been proposed in the literature, in the following, we adapt a recent technique, CORrelation ALignment (CORAL)~\cite{Sun15:CORAL}, to align the distributions of the features from the Android and iOS datasets. 
CORAL minimizes the shift between two domains, referred to as {\em source} and {\em target} domains,  by aligning the second-order statistics of source and target distributions. While CORAL is extremely simple, it has been shown to be effective,  efficient, and achieve similar performance as more complex approaches~\cite{Sun16a:DeepCORAL}. 

We next present three approaches of domain adaptation for our dataset. The first approach treats Android dataset as the source, and the iOS dataset as the target, which we refer to  as {\em Android-transformed}. The second method switches the roles of Android and iOS datasets, and we refer to it as {\em iOS-transformed}. In the third method, we transform both Android and iOS datasets to a common feature space, which we refer to as {\em dual-transformed}. 
All three approaches need balanced datasets for Android and iOS datasets. We therefore first describe data balancing, and then describe the three domain adaptation approaches.

{\bf Data balancing.} The Android dataset has much less samples than the iOS dataset (202 vs. 462). In addition, the Android dataset has 39 improved and 163 not-improved samples, significantly more imbalanced compared to the relatively balanced samples for iOS (194 improved and 268 not-improved). To balance the Android dataset, we first upsampled the 39 improved samples by a factor of 4 by duplication to form 156 improved samples. This upsampling also increased the Android dataset to a total of 319 samples, which is still significantly less than the total number of samples in the iOS dataset. We then uniformly upsampled the Android dataset by a factor of 1.4 (again by duplication) to increase it to 447 samples. We then apply domain adaptation to the 447 Android samples and 462 iOS samples.

\input{CORAL-res}

\emph{\noindent{\bf Android-transformed.}} As mentioned earlier, this approach treats Android dataset as the source and iOS dataset as the target. It follows CORAL domain adaptation directly. Let $X_s$ and $X_t$ denote the source and target domain feature matrices, respectively, where $X_s$ is of size $N_s \times D$ and $X_t$ is of size $N_t \times D$, where $N_s$ and $N_t$ are the number of samples for the source and target domains, respectively and $D$ is the dimension of the feature space. Let $\mu_s$ and $\mu_t$ denote the feature vector mean for the source and target domain features, respectively. For both source and target domains, we normalize the features (i.e., subtracting the mean value for each feature so that for the mean of the normalized feature values is zero). After feature normalization, we obtain the feature covariance matrices, denoted as $C_s$ and $C_t$ for the source and target domain, respectively. To minimize the distance between the second-order statistics (covariance) of the source and target features, CORAL applies a linear transformation $A$ to the original source features and uses the Frobenius norm as the matrix distance metric. Specifically, the objective function is 
\[
\mbox{minimize}_A  \quad \lVert A^T C_s A - C_t \rVert_F ^2,
\]
where $\lVert \cdot \rVert_F ^2$ represents the matrix Frobenius norm.
Once $A$ is determined, the transformed feature of the source is $C_s A + \mu_t$, which will be used together with the target dataset to train models for classification (see \S~\ref{sec:class}). 

Fig.~\ref{fig:features-CORAL-Android-transform} shows the distributions of the eight location features. For each feature, it compares the distributions of the original and transformed Android features, and the original iOS feature. We see that each transformed Android feature indeed exhibits greater similarity to the corresponding iOS feature than the original Android feature.

\emph{\noindent{\bf iOS-transformed.}} This approach is similar to Android-transformed approach except that it treats iOS feature space as the source, and transforms it to resemble the Android feature space. We observe that the various transformed iOS features are indeed closer to their corresponding Android features (figure omitted).

\emph{\noindent{\bf Dual-transformed.}} In this approach, we extend CORAL to transform Android and iOS feature spaces to a common space. Specifically, let $X_{a,i}$ denote the combined feature matrices for Android and iOS, which is of dimension $N_a+N_i$, where $N_a$ and $N_i$ are the number of samples for the Android and iOS datasets, respectively. Let $\mu_{a,i}$ denote the feature vector mean of $X_{a,i}$. We then apply similar normalization process to $X_{a,i}$, and denote the covariance matrix after the normalization as $C_{a,i}$. Similarly, let $\mu_{a}$ and $\mu_{i}$ denote the feature vector mean for the Android and iOS feature space, respectively. Let $C_{a}$ and $C_{i}$ denote their respective covariance matrix after the normalization process. To find transformation matrix, $A_a$, for the Android feature space, we solve the following minimization problem 
\[
\mbox{minimize}_{A_a}  \quad \lVert A_a^T C_a A_a - C_{a,i} \rVert_F ^2,
\] 
Similarly, we find transformation matrix $A_i$ for the iOS feature space as
\[
\mbox{minimize}_{A_i}  \quad \lVert A_i^T C_i A_i - C_{a,i} \rVert_F ^2,
\]
Once $A_a$  and $A_i$ are determined, the transformed features for Android and iOS are  $C_{a,i} A_a + \mu_{a,i}$ and  $C_{a,i} A_i + \mu_{a,i}$, respectively. 
Fig.~\ref{fig:features-CORAL-Both-transform} plots the distributions of the transformed Android and iOS features together with their original features. We see that, for each feature, the distributions of the transformed features are indeed closer than those of the original features, demonstrating the effectiveness of this domain adaptation approach.

%% file: CORAL-res.tex
%%%% Android_TF%%%%%%%%%%%%%%%%%%%%%%%%%%%%%%%%%%%%%%

\begin{figure}[ht]
    \centering
    \begin{subfigure}{0.25\textwidth}
        \includegraphics[width=\linewidth]{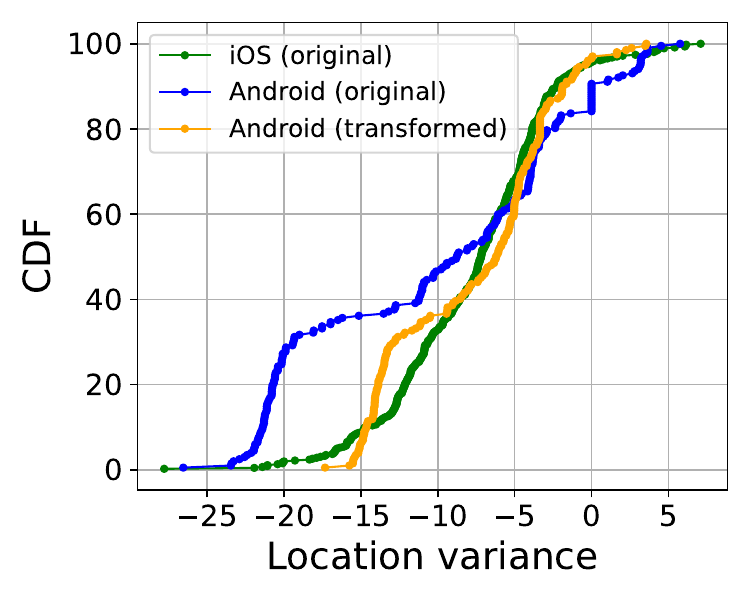}
        \caption{}
        % \label{fig:figure_location_variance_android}
    \end{subfigure}
    \hspace{-0.1in}
    \begin{subfigure}{0.25\textwidth}
        \includegraphics[width=\linewidth]{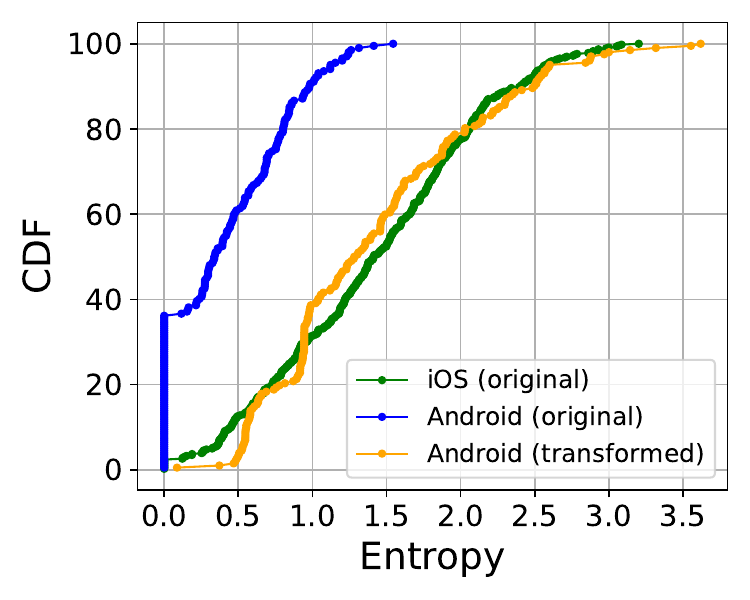}
        \caption{}
        % \label{fig:figure2}
    \end{subfigure}
    \hspace{-0.1in}
    \begin{subfigure}{0.25\textwidth}
        \includegraphics[width=\linewidth]{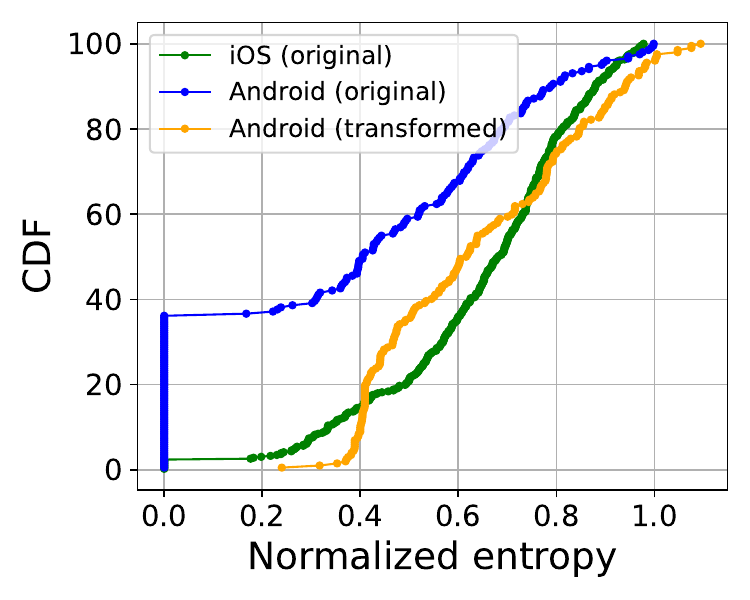}
        \caption{}
        % \label{fig:figure3}
    \end{subfigure}
    \hspace{-0.1in}
    \begin{subfigure}{0.25\textwidth}
        \includegraphics[width=\linewidth]{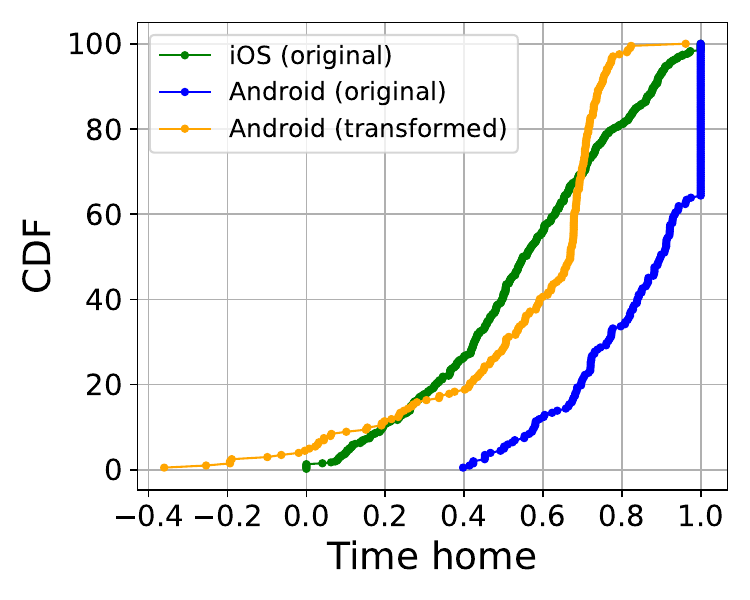}
        \caption{}
        % \label{fig:figure4}
    \end{subfigure}

    % \medskip
    \vspace{10pt} % add vertical space here

    \begin{subfigure}{0.25\textwidth}
        \includegraphics[width=\linewidth]{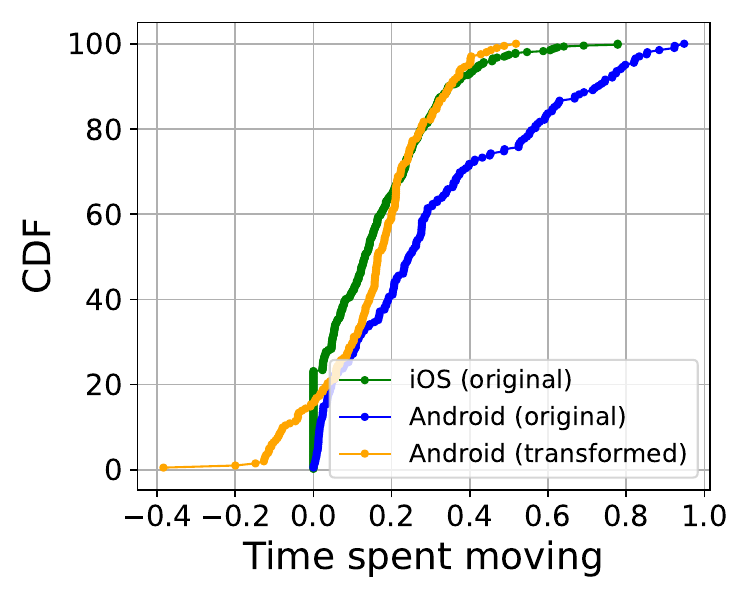}
        \caption{}
        % \label{fig:figure5}
    \end{subfigure}
    \hspace{-0.1in}
    \begin{subfigure}{0.25\textwidth}
        \includegraphics[width=\linewidth]{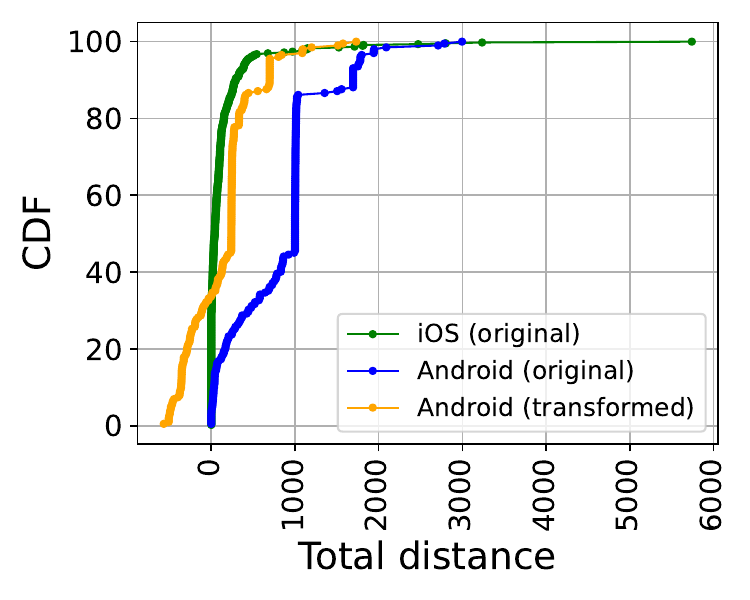}
        \caption{}
        % \caption{Total Distance (in Km)}
        \label{fig:figure6}
    \end{subfigure}
    \hspace{-0.1in}
    \begin{subfigure}{0.25\textwidth}
        \includegraphics[width=\linewidth]{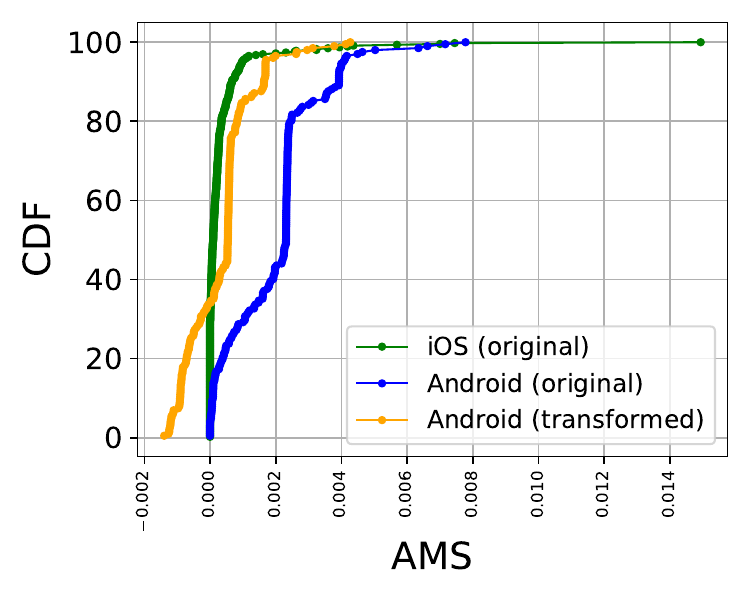}
        \caption{}
        % \caption{AMS( km/hr) }
        % \label{fig:figure7}
    \end{subfigure}
    \hspace{-0.1in}
    \begin{subfigure}{0.25\textwidth}
        \includegraphics[width=\linewidth]{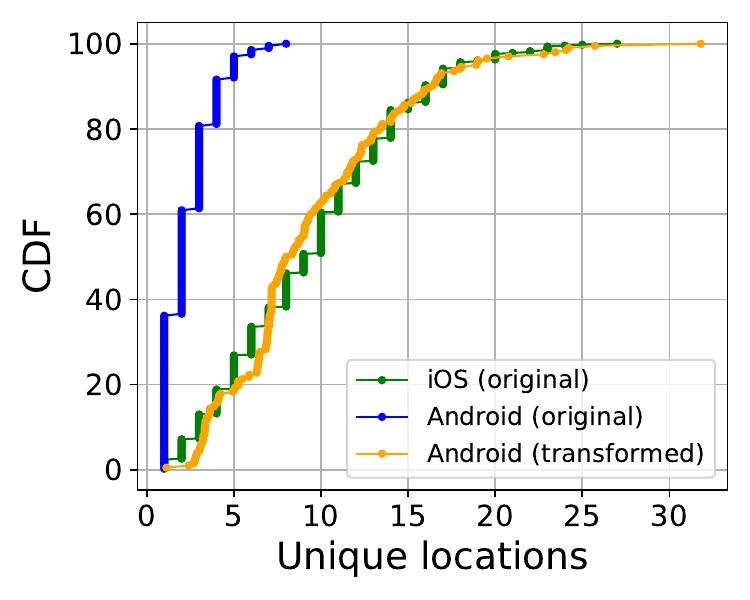}
        \caption{}
        % \label{fig:figure8}
    \end{subfigure}
    \caption{Android-transformed: distributions of location features; each plot shows the original iOS features, and the original and transformed Android features. }
    \label{fig:features-CORAL-Android-transform}
\end{figure}

%%%% Both_TF%%%%%%%%%%%%%%%%%%%%%%%%%%%%%%%%%%%%%%

\begin{figure}[ht]
    \centering
    \begin{subfigure}{0.25\textwidth}
        \includegraphics[width=\linewidth]{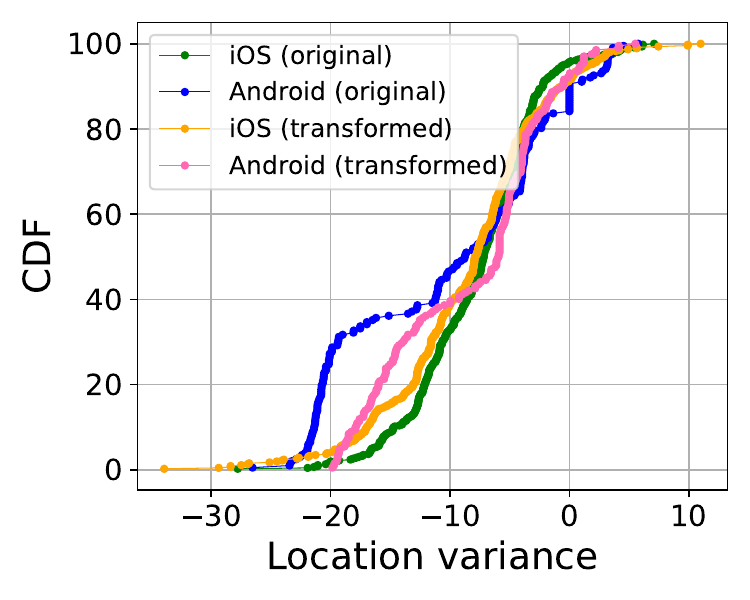}
        \caption{}
        % \label{fig:figure_location_variance_BothTF}
    \end{subfigure}
    \hspace{-0.1in}
    \begin{subfigure}{0.25\textwidth}
        \includegraphics[width=\linewidth]{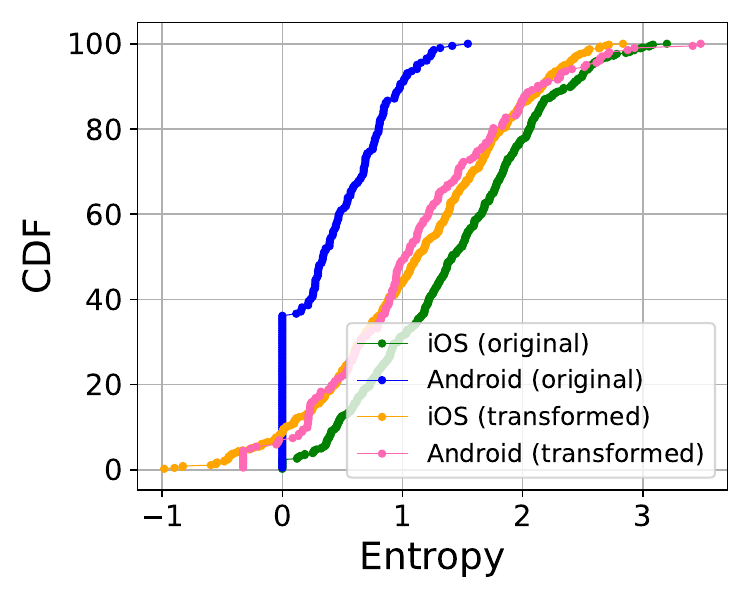}
        \caption{}
        % \label{fig:figure2}
    \end{subfigure}
    \hspace{-0.1in}
    \begin{subfigure}{0.25\textwidth}
        \includegraphics[width=\linewidth]{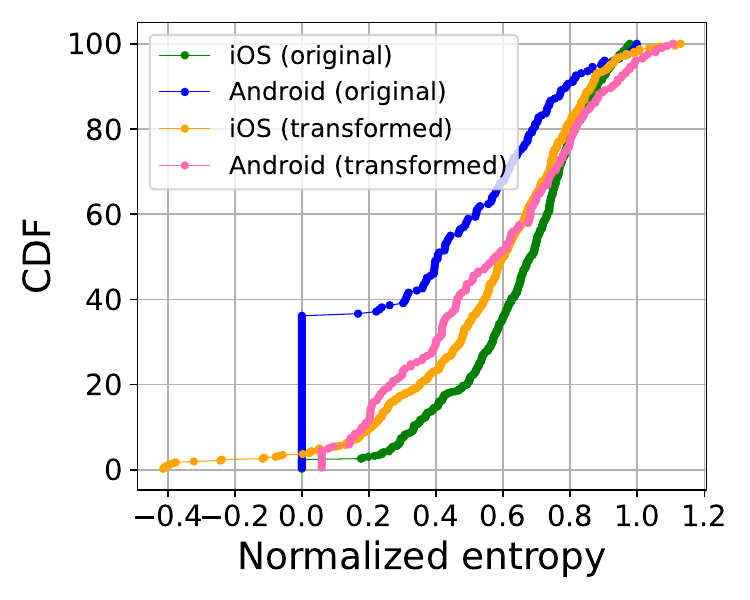}
        \caption{}
        % \label{fig:figure3}
    \end{subfigure}
    \hspace{-0.1in}
    \begin{subfigure}{0.25\textwidth}
        \includegraphics[width=\linewidth]{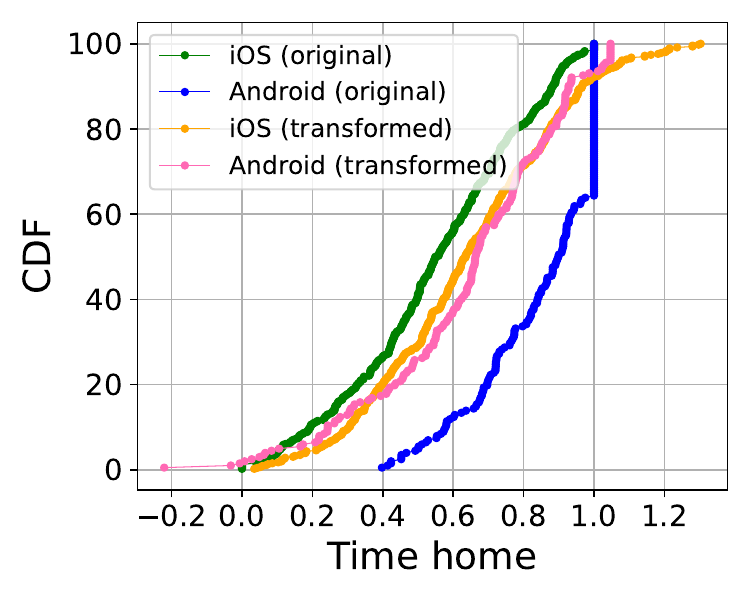}
        \caption{}
        % \label{fig:figure4}
    \end{subfigure}

    % \medskip
    \vspace{10pt} % add vertical space here

    \begin{subfigure}{0.25\textwidth}
        \includegraphics[width=\linewidth]{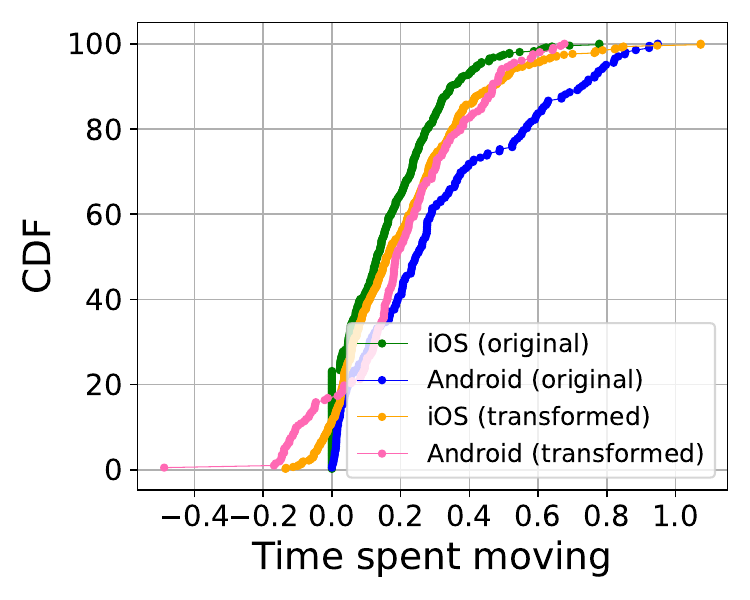}
        \caption{}
        % \label{fig:figure5}
    \end{subfigure}
    \hspace{-0.1in}
    \begin{subfigure}{0.25\textwidth}
        \includegraphics[width=\linewidth]{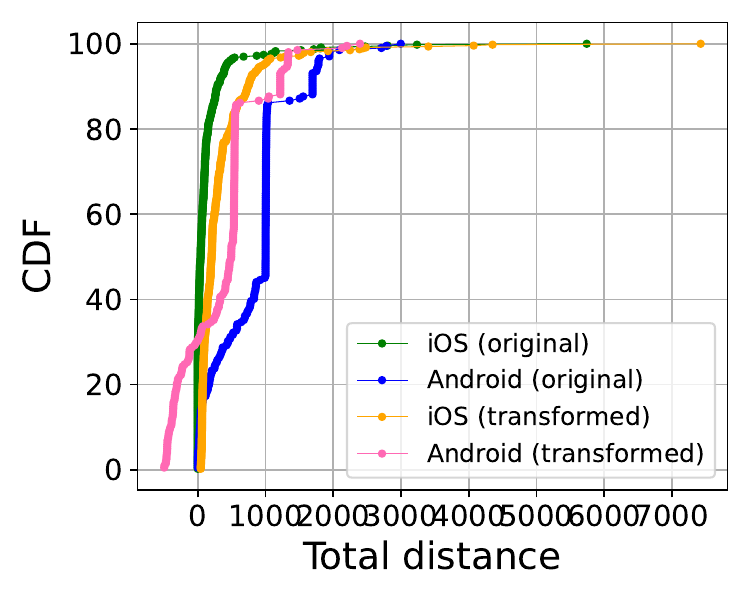}
        \caption{}
        % \caption{Total Distance (in Km)}
        % \label{fig:figure6}
    \end{subfigure}
    \hspace{-0.1in}
    \begin{subfigure}{0.25\textwidth}
        \includegraphics[width=\linewidth]{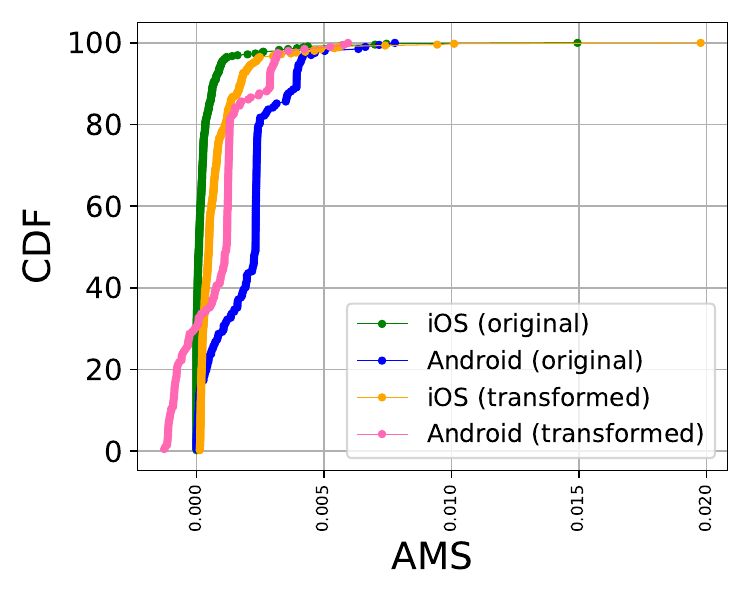}
        \caption{}
        % \caption{AMS( km/hr) }
        % \label{fig:figure7}
    \end{subfigure}
    \hspace{-0.1in}
    \begin{subfigure}{0.25\textwidth}
        \includegraphics[width=\linewidth]{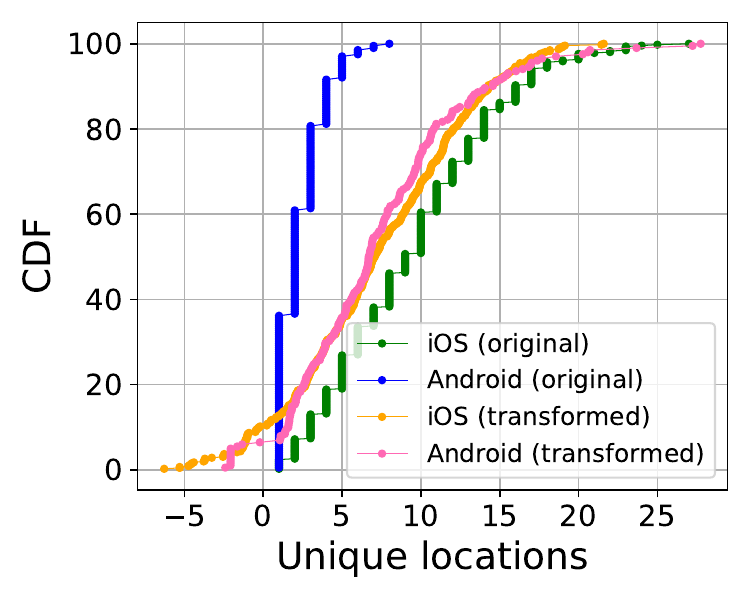}
        \caption{}
        % \label{fig:figure8}
    \end{subfigure}
    \caption{Dual-transformed: distributions of location features; each plot shows the original iOS and Android features, and transformed iOS and Android features. }
    \label{fig:features-CORAL-Both-transform}
\end{figure}

%% file: correlation-analysis.tex
%%%%%%%%%%%%%%%%%%%%  correlation %%%%%%%

\subsection{Correlation Analysis}

We obtain Pearson correlation coefficients between location features and self-reported QIDS scores for each QIDS interval for both the original datasets and the transformed datasets (using all three approaches of Android-transformed, iOS-transformed, and dual-transformed).  In each case, we find features that are correlated with QIDS scores, indicating that location features can be used to predict depression status. We next only present the correlation results for the dual-transformed approach. As we shall see in \S\ref{sec:class}, this approach leads to the best prediction results.

%==========================================================

\begin{table}[ht]
\caption{Correlation results for dual-transformed iOS and Android features.} 
\label{tab:Feature Corelation BothTF merged- Pearson's}
\begin{tabular}{|l|ll|ll|ll|}
\hline
                              & \multicolumn{2}{c|}{\textbf{All}}                          & \multicolumn{2}{c|}{\textbf{Improved}}                     & \multicolumn{2}{c|}{\textbf{Not-improved}}                  \\ \hline
\textbf{Features}             & \multicolumn{1}{l|}{\textbf{r-values}} & \textbf{p-values} & \multicolumn{1}{l|}{\textbf{r-values}} & \textbf{p-values} & \multicolumn{1}{l|}{\textbf{r-values}} & \textbf{p-values} \\ \hline
\textbf{location   variance}  & \multicolumn{1}{l|}{-0.08}             & 0.04              & \multicolumn{1}{l|}{-0.02}             & 0.73              & \multicolumn{1}{l|}{-0.13}             & 0.01              \\ \hline
\textbf{time spent   moving}  & \multicolumn{1}{l|}{-0.04}             & 0.26              & \multicolumn{1}{l|}{-0.01}             & 0.84              & \multicolumn{1}{l|}{-0.10}             & 0.04              \\ \hline
\textbf{total distance}       & \multicolumn{1}{l|}{0.05}              & 0.24              & \multicolumn{1}{l|}{0.06}              & 0.34              & \multicolumn{1}{l|}{0.06}              & 0.18              \\ \hline
\textbf{AMS}                  & \multicolumn{1}{l|}{0.04}              & 0.33              & \multicolumn{1}{l|}{0.06}              & 0.38              & \multicolumn{1}{l|}{0.16}              & 0.05              \\ \hline
\textbf{unique   locations}   & \multicolumn{1}{l|}{-0.17}             & $10^{-5}$              & \multicolumn{1}{l|}{-0.03}             & 0.06              & \multicolumn{1}{l|}{-0.19}             & $6 \times 10^{-5}$              \\ \hline
\textbf{entropy}              & \multicolumn{1}{l|}{-0.17}             & $10^{-5}$              & \multicolumn{1}{l|}{-0.04}             & 0.06              & \multicolumn{1}{l|}{-0.19}             & $10^{-4}$               \\ \hline
\textbf{normalized   entropy} & \multicolumn{1}{l|}{-0.14}             & $2 \times 10^{-4}$              & \multicolumn{1}{l|}{-0.08}             & 0.06              & \multicolumn{1}{l|}{-0.14}             & $3 \times 10^{-3}$              \\ \hline
\textbf{time home}            & \multicolumn{1}{l|}{0.08}              & 0.04              & \multicolumn{1}{l|}{0.01}              & 0.37              & \multicolumn{1}{l|}{0.10}              & 0.04              \\ \hline
\end{tabular}
\end{table}

Table~\ref{tab:Feature Corelation BothTF merged- Pearson's} shows Pearson correlation coefficients between location features in a week and the corresponding self-reported QIDS scores for the dual-transformed data (including both transformed Android and iOS data). It shows the correlation for three categories, all the samples, improved samples only, and not-improved samples only.  We see that for all the samples, five features, location variance, unique locations, entropy, normalized entropy and time spent at home, are significantly correlated with QIDs scores. Specifically, the first four features are negatively correlated with QIDS score, while the last (time spent at home) is positively correlated with QIDS score. This is consistent with findings in earlier studies~\cite{cacioppo2010perceived,sanders2000relationship,martinsen1990benefits,saeb2015mobile,lane2014bewell,Lathia:2013:OSS:2494091.2497345,Chow17:mobilesensing} 
%Farhan16:depression,Yue18:fusion,Ware2018:WifiMetadata,
that depression is often linked with social isolation. Similar observations hold for not-improved samples, and an additional feature, time spent in moving, is also negatively correlated with QIDS score. For improved samples, only three features, unique locations, entropy, and normalized entropy, are significantly correlated with QIDS score, with low $r$-values.  

%% file: predict.tex
\section{ Predicting Depression treatment Outcome} \label{sec:class}

In this section, we develop machine learning based classification models that predict the improvement or lack of improvement of depression symptoms using the location features described in \S\ref{sec:corr}. In the following, we first describe the classification  methodology and then the results.

\subsection{Classification Methodology} \label{sec:xgboost}

The classification is for each QIDS interval, which contains the QIDS score and the location features extracted from the location data collected in the interval (a week). In addition, we further consider {\em location baseline}, which represents location-related behavior at the beginning of the treatment and is obtained using the location data collected in the week right after the enrollment. 

The clinical ground truth, i.e., CGI-I score assessed by the study clinician,  served as the label for improvement status (see \S\ref{sec:data-collection} and \S\ref{sec:corr}). We explore four scenarios, all using combined iOS and Android datasets: (i) {\em non-transformed}, i.e., simply concatenating the Android and iOS datasets together, (ii) {\em Android-transformed}, i.e., treating Android dataset as source, and iOS dataset as target, and transform Android dataset to be in the feature space of the iOS dataset, (iii) {\em iOS-transformed}, which differs from Android-transformed in that iOS dataset is transformed to the feature space of the Android dataset, and (iv) {\em dual-transformed}, which transforms Android and iOS datasets jointly  into a common feature space. The latter three scenarios were obtained using domain adaptation; see \S\ref{sec:domain-adaptation}.

For all the above four scenarios, we used leave-one-user-out cross validation procedure,  i.e., we used $N-1$ users' data as the training set (which includes a mixture of data from both iPhone and Android users), and one user's data (either iOS or Android) as the testing set, where $N$ is the number of users (i.e., 66 for our data analysis). We repeated the above procedure $N$ times to obtain the validation results for all the users.

\emph{\noindent{\bf Classification Algorithms.}} We explored two classification algorithms: Support Vector Machine (SVM)~\cite{boser1992training,cortes1995support} with radial basis function (RBF)  kernel~\cite{CC01a:libSVM} and XGBoost~\cite{chen2016xgboost}, and compare their prediction accuracy. Each algorithm involves tuning multiple hyperparameters. We chose the hyperparameters that gave the best validation $F_1$ score, which is the harmonic mean of precision and recall, i.e., $2  {(\textrm{precision} \times \textrm{recall})}/{(\textrm{precision} + \textrm{recall})}$.  $F_1$ score ranges from $0$ to $1$, and the higher, the better. We next briefly describe the hyperparameters in each algorithm and our approach for feature selection.

\begin{itemize}
    \item SVM with RBF kernel has two hyperparameters, the cost parameter $C$ and the parameter $\gamma$ of the radial basis functions. We varied  $C$ and $\gamma$ both in $2^{-15}, 2^{-14}, \ldots, 2^{14}, 2^{15}$. We used SVM recursive feature elimination (SVM-RFE)~\cite{guyon2002gene,rakotomamonjy2003variable,yan2015feature} for feature selection. Specifically, for a set of $n$ features, for each pair of values for $C$ and $\gamma$,  SVM-RFE provided a ranking of the features, from the most important to the least important. After that, for each feature, we obtained its average ranking across all the combinations of  $C$ and $\gamma$ values, leading to a complete order of the features. We then varied the number of features, $m$, from 2 to $n$. For a given $m$, the top $m$ features were used to choose the parameters, $C$ and $\gamma$, to maximize $F_1$ score based on the leave-one-user-out cross validation procedure as described above. The set of top $m$ features that provides the highest $F_1$ score is chosen as the best set of features.

\item XGBoost has several hyperparameters. We varied them in the following ranges: the maximum depth of a tree from 2 to 10, the minimum child weight (i.e., the minimum sum of weights of all observations required in a child of a tree, which was used to control over-fitting) from 1 to 6, the fraction of observations to be randomly sampled for each tree and the fraction of features to be randomly sampled for each tree from 0.1 to 5, and the gamma value (i.e., the minimum loss reduction required to make a further partition on a leaf node of a tree) 0.1 to 7, and the learning rate was varies from 0.1 to 0.3. For a given setting, we first ran XGBoost using all the features in the setting to obtain the best result (i.e., the highest $F_1$ score based on leave-one-user-out cross validation) and ranked the importance of the features. We then chose the top $m$ features (based on the importance scores), and varied $m$ from 2 to the total number of features. The set of $m$ features in combination with parameter tuning of XGBoost that provided the highest $F_1$ score was chosen as the best set of features.

\end{itemize}

\input{predict-results}

%% file: predict-results.tex
\subsection{Prediction Results}

In the following, we present prediction results. Our goal is to answer the following questions: (i) for the overall dataset (including both Android and iOS datasets), does domain adaptation lead to better prediction compared to no domain adaptation? (ii) for the dataset from one platform (Android or iOS), does a model trained using the combined dataset (either with or without domain adaptation) lead to better results than using the dataset alone? To answer the above two questions, we evaluate various settings, first examining the overall prediction accuracy for the combined data (\S\ref{sec:predict-result-overall}), and then the prediction accuracy for iOS and Android datasets separately (\S\ref{sec:res-ios} and \S\ref{sec:res-Android}).

For each setting, we explore the following six scenarios based on the features that are used for prediction:
\begin{itemize}
    \item {\bf QIDS + QIDS baseline}, which uses the current QIDS score in the QIDS interval and the baseline QIDS score. 
    
    \item {\bf Location}, which uses the 8 location features (see \S\ref{sec:corr}) obtained from the current QIDS interval.
        
    \item {\bf Location + Location baseline}, where location baseline includes the 8 features extracted using the data in the first week after the enrollment. 

    \item {\bf Location + QIDS baseline}, which uses 9 features,
    %as input, 
    including the  baseline QIDS score, and the 8 location features for the QIDS interval. 

    \item {\bf Location + QIDS baseline + Location baseline}, which uses 17 features as input, including  the 8 location features  for the current QIDS interval, the  baseline QIDS score, and the 8 location baseline features.

    \item {\bf All}, which uses QIDS + QIDS Baseline + location + location baseline, a total of 18 features as input, including 2 QIDS related features (current and baseline QIDS scores),  8 location features for the current QIDS interval, and 8 baseline location features.
\end{itemize}  
For all the settings, we used feature extraction described for SVM and XGBoost (see \S\ref{sec:xgboost}) to select the best set of features, which tends to be (significantly) less than the total number of  features. 

In the above, the setting using QIDS + QIDS baseline serves as a baseline setting since it represents the current practice of using self-reported questionnaire to keep track of depression symptom improvement status. The two settings, Location and Location + Location baseline, only leverage automatically collected sensory data, requiring no user interaction, and hence lead to the least amount of burden to  participants. The two settings that involve QIDS baseline and location sensory data, i.e.,  Location + QIDS baseline and  Location + QIDS baseline + Location baseline, lead to little burden to participants since baseline questionnaire score is often collected routinely before treatment starts. The last setting that uses all the QIDS and location sensory data serves to quantify how much benefits we obtain by using both types of data.

\subsubsection{Overall Prediction Results} \label{sec:predict-result-overall}
\input{predict-result-overall_combined}

Fig.~\ref{fig:combinedresult} shows the overall prediction results for the combined iOS and Android dataset using SVM and XGBoost. The results for the four forms of domain adaptation, non-transformed, Android-transformed, iOS-transformed, and dual-transformed, are shown in the figure. %Figures~\ref{fig:combinedresult}(a) and (b) show the results when using SVM and XGBoost, respectively, where 
In both Fig.~\ref{fig:combinedresult}a and Fig.~\ref{fig:combinedresult}b, the comparison baseline (i.e., the result obtained using QIDS + QIDS baseline) is marked as a dotted horizontal line. 
The rest of the settings are divided into multiple groups based on the features that are used, and each group includes the results for the four different forms of domain adaptation, represented in bars with different colors. 

When using QIDS + QIDS baseline, the $F_1$ score for SVM is 0.68, similar to that for  XGBoost (0.69). For the rest of the settings that involve location data, we see increasingly higher predicted $F_1$ score in the order of non-transformed, Android-transformed, iOS-transformed, and dual-transformed. The better results with transformation 
compared to those with non-transformed data indicate that domain adaptation is indeed helpful in aligning the feature spaces of the Android and iOS datasets, which leads to better prediction results. The results of dual-transformed are better than transforming a single dataset, which indicates that transforming both datasets jointly to a common feature space leads to more benefits.  The setting of using all the features leads to similar $F_1$ score as that obtained by using QIDS + QIDS baseline, indicating that, when combining QIDS and location sensory features,  QIDS features play more dominant roles in prediction accuracy than location sensory features. For the rest of the settings,
we see SVM achieves better results than XGBoost. In the following, we only present the results for these settings using SVM.    

When using SVM (see Fig.~\ref{fig:combinedresult}a), 
we see that Location + Location baseline and Location + QIDS baseline lead to higher $F_1$ score than using current location features only. This implies that location baseline and QIDS baseline are complementary to the current location features, maybe because they provide insights into individual variation, which allows the models to be more
customized to each individual. Similar observations are made in~\cite{Shende2023:PredictingSymptom}, which showed that combining sleep baseline and current sleep features are helpful in predicting depression improvement status. 
Combining both QIDS baseline and location baseline features with current location features leads to even better results. The best $F_1$ score is 0.65, achieved under dual-transformation for Location + QIDS baseline + Location baseline, which is very close to the $F_1$ score (0.68) obtained using QIDS and QIDS baseline. Since this setting requires only participant to fill in QIDS baseline questionnaire, which is much less burdensome than requiring participants to fill in QIDS questionnaire periodically, it represents a promising direction for predicting depression treatment status. Note that even though non-transformed case also leads to better results in this setting compared to the previous three settings (i.e., Location, Location + Location baseline, and Location + QIDS baseline), its $F_1$ score is only 0.60, significantly lower than 0.65 achieved by the dual-transformed case.

\input{predict-results-iOS}

\input{predict-results-Android}

%% file: predict-result-overall_combined.tex
\begin{figure}[t]
    \centering   
    \begin{subfigure}[b]{0.6\textwidth}
        \centering
        \includegraphics[width=\textwidth]{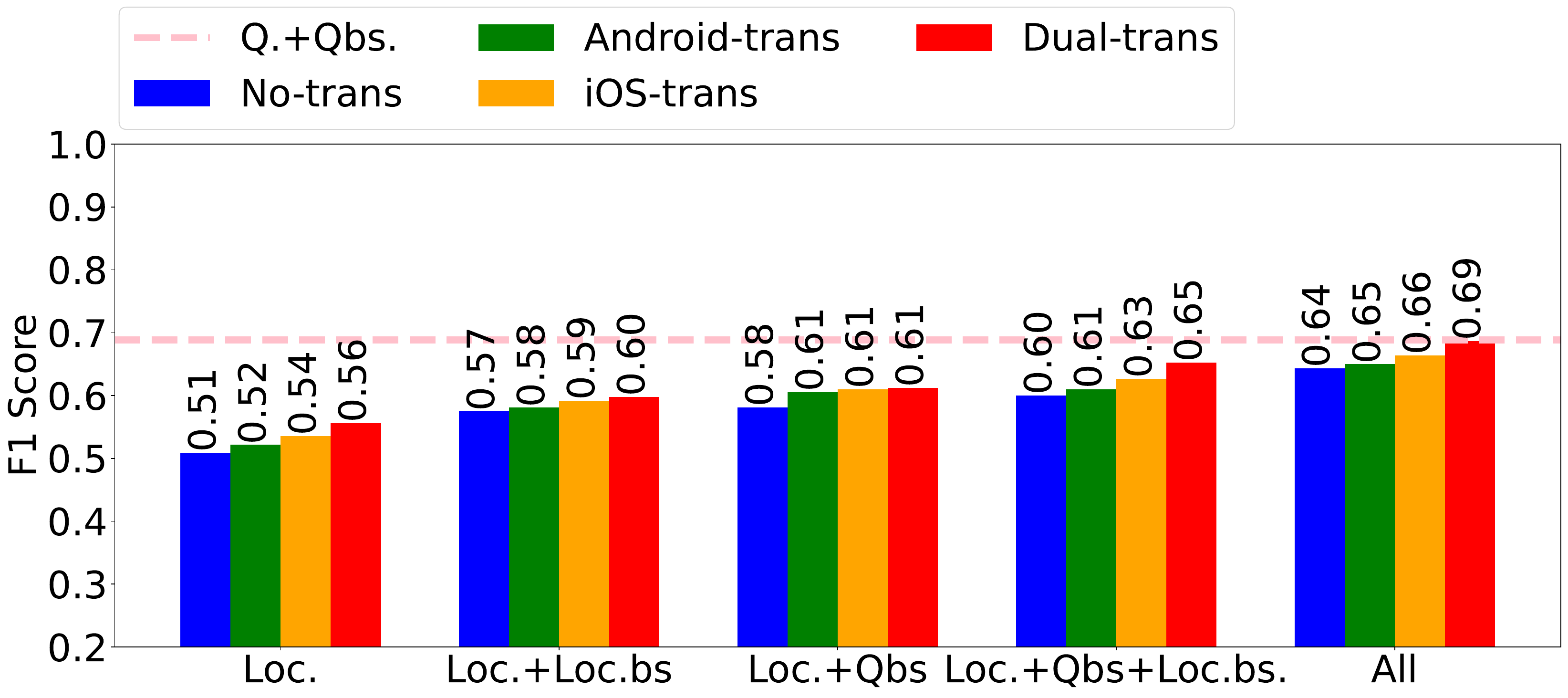}
        \caption{Overall results (SVM).}
    \end{subfigure}%
    \hfill \\
    \vspace{0.5cm}
    \begin{subfigure}[b]{0.6\textwidth}
        \centering
        \includegraphics[width=\textwidth]{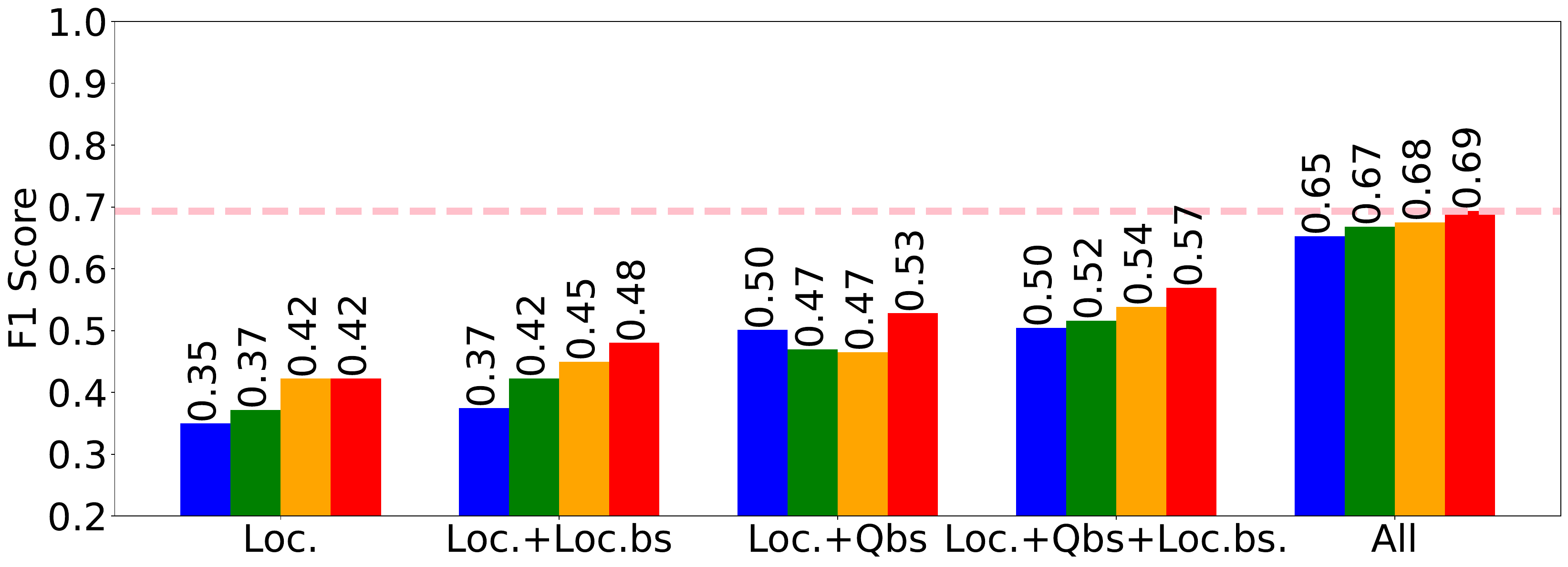}
        \caption{Overall results (XGBoost).}
    \end{subfigure}    
    \caption{Prediction results for all the users when varying features and domain adaptation methods, in comparison to no domain adaptation.  %\soumya{(SVM: Q. + Qbs. = 0.68), (XGB: Q. + Qbs. = 0.69)} 
    } 
    \label{fig:combinedresult}
\end{figure}

%% file: predict-results-iOS.tex
%%%%%%%%%%%%%   IOS results %%%%%%%%%%%%%%%
\begin{figure}[h]
    \centering   
    \begin{subfigure}[b]{0.7\textwidth}
        \centering
        \includegraphics[width=\textwidth]{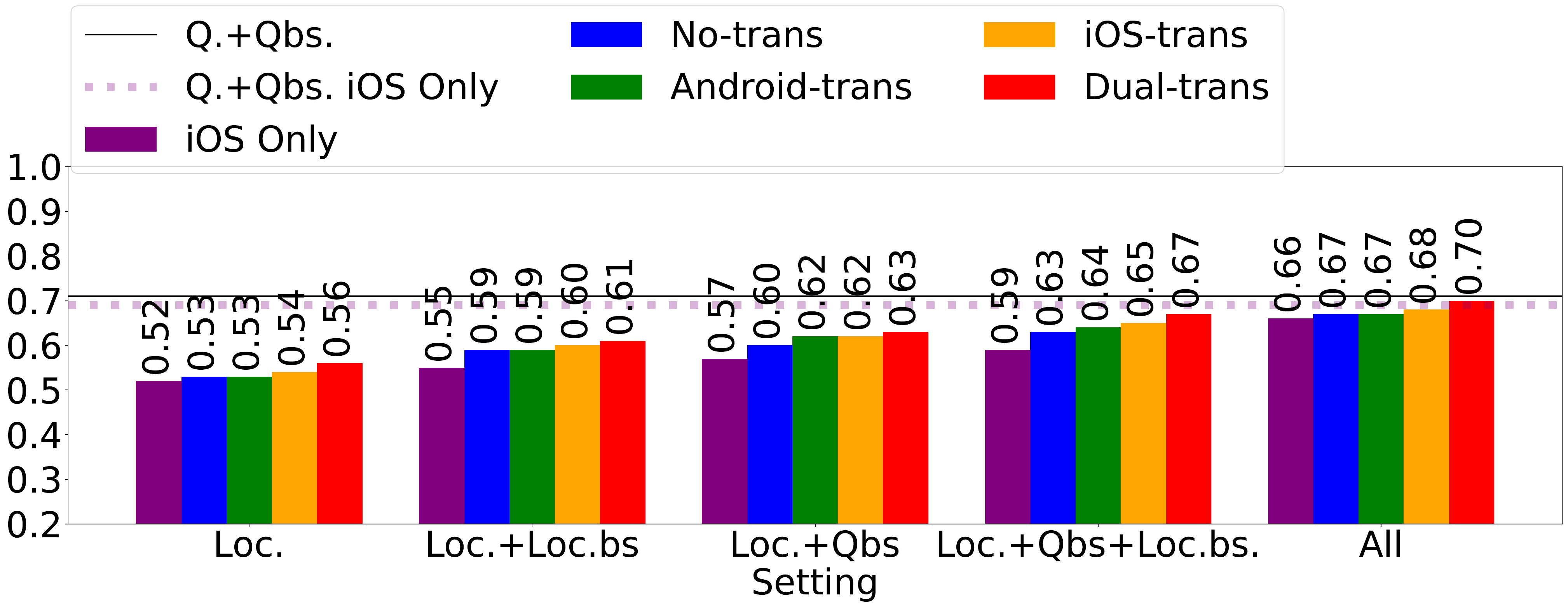}
        \caption{iOS results (SVM).
        }
        
    \end{subfigure}%
    \hfill \\
    \vspace{0.5cm}
    \begin{subfigure}[b]{0.7\textwidth}
        \centering
        \includegraphics[width=\textwidth]{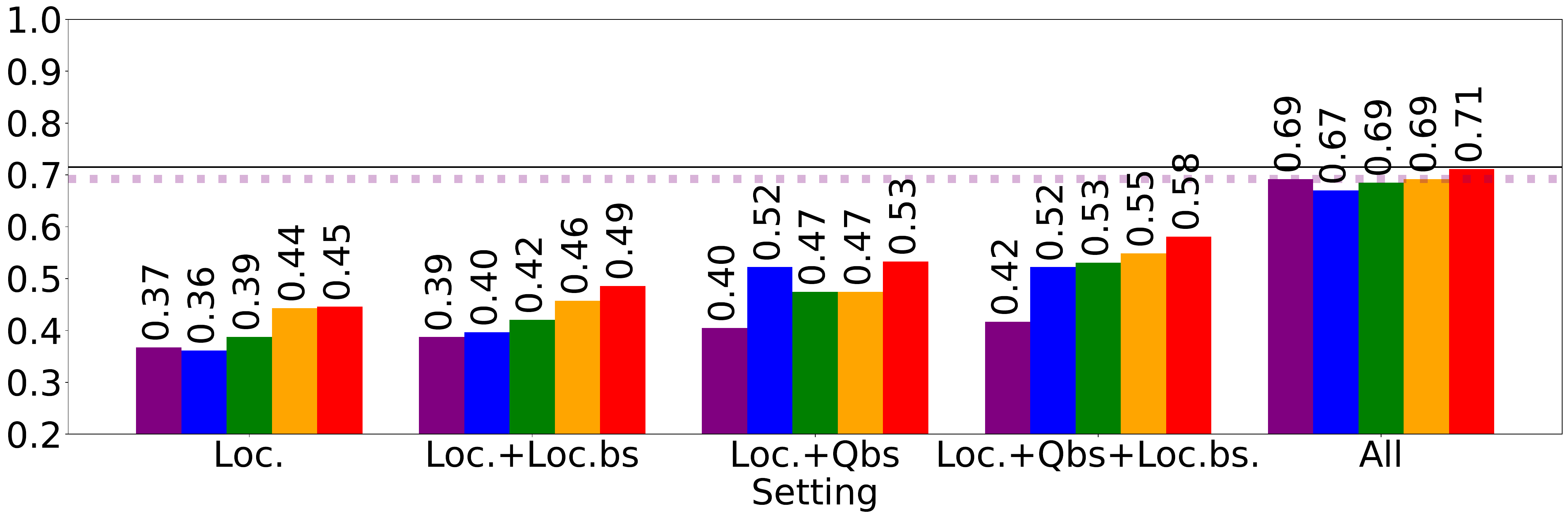}
        \caption{iOS results (XGBoost). 
        }
    \end{subfigure}    
    \caption{Prediction results for iOS users when varying features and domain adaptation methods, in comparison to prediction results with no domain adaptation and those using iOS data alone. } 
    \label{fig:iosresult}
\end{figure}

\subsubsection{Prediction Results for iOS Dataset} \label{sec:res-ios}
We next present the prediction results for the iOS dataset. One straightforward approach is using the iOS dataset alone 
for prediction. Another approach is using a larger training dataset by combining Android and iOS dataset together. We again consider the four forms of domain adaptation as in \S\ref{sec:predict-result-overall}. 
For each form of domain adaptation, we isolated the results pertaining to iOS users to obtain $F_1$ score accordingly.

Fig.~\ref{fig:iosresult} shows the results for SVM and XGBoost in a similar way as shown in Fig.~\ref{fig:combinedresult}. Additionally, for comparison, prediction results are obtained from the original iOS dataset %without any transformation 
(i.e., the purple bars).  In Fig.~\ref{fig:iosresult}, the two horizontal lines in each sub-plot represent the baseline prediction results. Specifically, the solid black line represents the prediction using the combined dataset (i.e., QIDS and QIDS baseline scores from both Android and iOS users); no domain adaptation is needed for this case since QIDS data does not depend on sensing platform.  The dotted purple line represents the prediction when using only QIDS and QIDS baseline from the iOS dataset. 

We next compare prediction results when using the combined dataset for training versus those using only the iOS dataset. As baseline prediction (i.e., when using QIDS and QIDS baseline), the $F1$ score using the combined dataset is similar for both classification algorithms (0.71 for SVM and 0.72 for  XGBoost). When using only iOS dataset, both algorithms obtain $F1$ score of 0.69. As expected, using the combined dataset for training leads to better prediction than that using iOS dataset alone. 

We now consider the various settings that involve location sensory data. For SVM, we see that for all the five settings involving location data, using iOS dataset alone leads to worse prediction than the four forms of transformation. For the four ways of transformation, we again observe increasingly better results in the order of non-transformed, Android-transformed, iOS-transformed, and dual-transformed. The results with transformation are again better than that without transformation, again indicating the benefits of domain adaptation.  When using iOS dataset alone, its performance is worse even than that of non-transformed. This is perhaps not surprising since combining Android and iOS datasets leads to a larger training set, which in general leads to better prediction. 

The above observations also hold in general when using XGBoost, except for several cases where using iOS data alone leads to slightly better results than using the combined dataset. On the other hand, dual-transformed case still leads to the best results in all the settings. 

In Fig.~\ref{fig:iosresult}, when QIDS score of the current interval is not used, we again see that the setting with Location + QIDS baseline + Location baseline for dual-transformed case leads to the highest $F_1$ score (0.67 when using SVM), which is close to the two horizontal lines (i.e., when using QIDS + QIDS baseline score).

%% file: predict-results-Android.tex
\subsubsection{Prediction Results for Android Dataset} \label{sec:res-Android}
\begin{figure}[t]
    \centering   
    \begin{subfigure}[b]{0.7\textwidth}
        \centering
        \includegraphics[width=\textwidth]{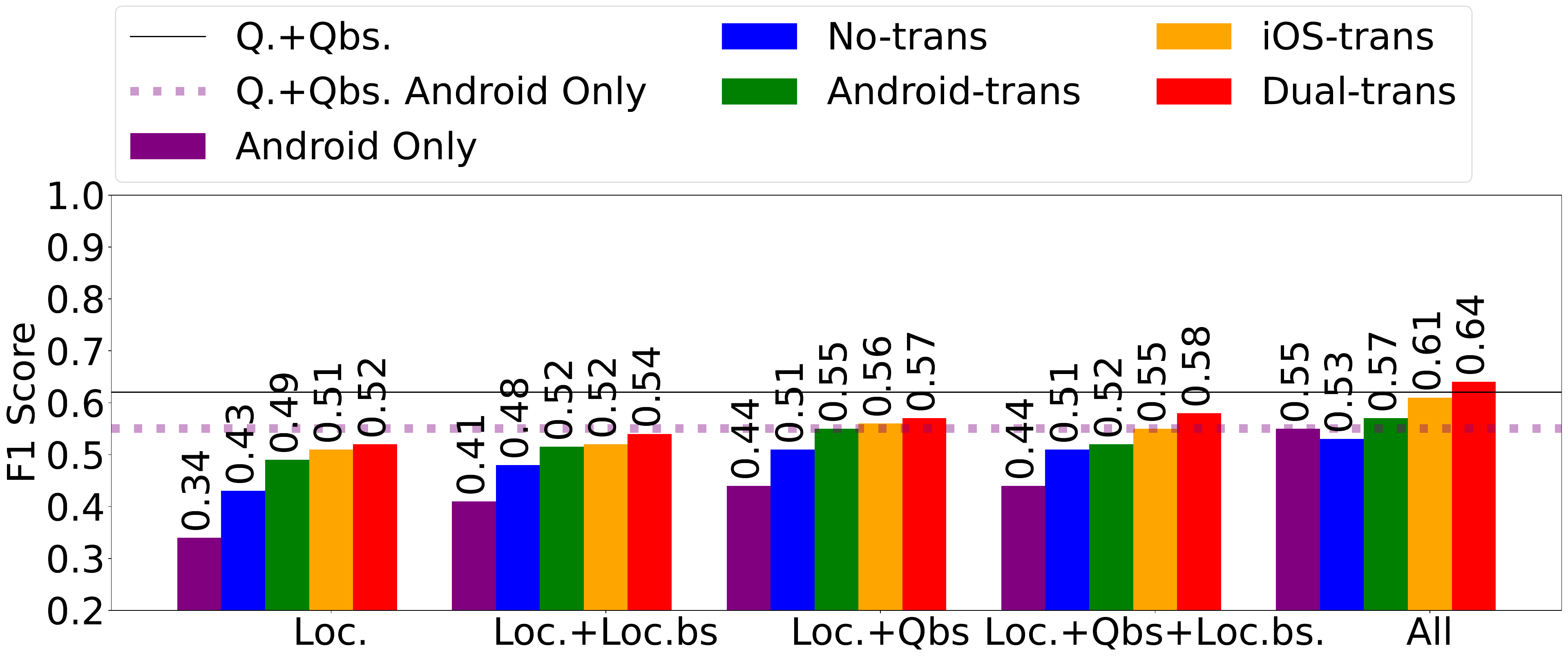}
        \caption{Android results (SVM). %
        }
    \end{subfigure}%
    \hfill \\
    \vspace{0.5 cm}
    \begin{subfigure}[b]{0.7\textwidth}
        \centering
        \includegraphics[width=\textwidth]{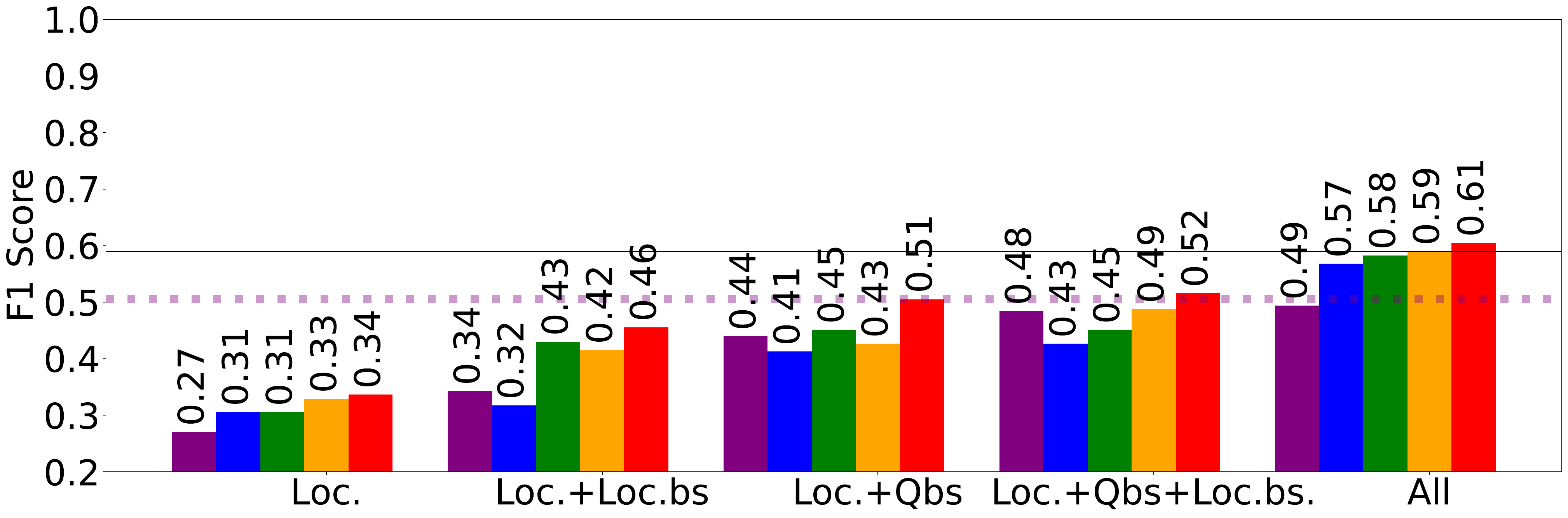}
        \caption{Android results (XGBoost). 
        }
    \end{subfigure}    
\caption{Prediction results for Android users  when varying features and domain adaptation methods, in comparison to prediction results with no domain adaptation and those using Android data alone.} 
\label{fig:androidresult}
\end{figure}

We next present the prediction results for the Android dataset. Again, our focus is on the impact of combined training set and domain adaptation on prediction accuracy. Fig.~\ref{fig:androidresult} presents the results for various settings, similar to those shown in Fig.~\ref{fig:iosresult} for the iOS dataset. We see that in general the results are worse than the overall results and the results for the iOS dataset. This might be because of the much smaller dataset for Android as well as the imbalance between improved and non-improved samples.  
When using QIDS and QIDS baseline, the $F_1$ score for the combined dataset is 0.61 and 0.59 for SVM and XGBoost, respectively (i.e., the solid black lines in Fig.~\ref{fig:androidresult}a and b); when using the Android dataset alone, the corresponding $F_1$ scores are 0.55 and 0.51.  

For the settings using location sensory data, consistent as our observations earlier, we see the worst performance when using the Android dataset alone, and the best results when using dual-transformed Android and iOS combined dataset. 

We next consider particularly the four settings that do not use periodic QIDS score. For these settings, the best $F_1$ score is 0.58 when using Location + Location baseline + QIDS baseline with dual-transformed data, only slightly lower than the result (0.61) when using QIDS + QIDS baseline with combined dataset, and higher than the result (0.55) when using QIDS + QIDS baseline with Android dataset alone. In contrast, the best $F_1$ score using Android dataset alone is only 0.44, and the best $F_1$ score using the combined non-transformed dataset is only 0.51.

%% file: discuss-SS.tex
\section{Discussion} \label{sec:discuss}

{\bf Main findings.}  
Our results demonstrate that applying domain adaptation %transforming 
to transform location data collected from different smartphone platforms into a common feature space is an effective approach to obtaining a larger dataset and training more accurate machine learning models. 
Specifically, when applying this approach, we obtain more accurate prediction of depression treatment outcome for all the users,  the iOS users, and Android users, compared to the results with no domain adaptation. In addition, among the three domain adaptation approaches that we explore, the dual-transformed approach that transforms Android and iOS together to a common feature space leads to the best results.

Our exploration in various scenarios also demonstrate that using location data along with QIDS baseline score can lead to $F_1$ score up to 0.67, close to the best $F_1$ score (0.69) obtain using self-reported scores (the current QIDS and baseline QIDS  scores). Our results therefore demonstrate that monitoring location sensory data provides a promising direction in predicting depression treatment, without burdening patients to fill in self-report questionnaires periodically.

{\bf limitations of our work.}  Our work uses a small sample size from 66 participants. Therefore, our results need to be validated using larger datasets. In addition, the dataset that we analyzed comes predominantly from female participants, which can bias the results. This gender imbalance is consistent with the findings that women are approximately twice as likely as men to be diagnosed with depression~\cite{Eid2019:Sexdifferences}, and are more likely to seek treatment (and participate in clinical studies)~\cite{Myrna2014:Treatment}. While we intend to recruit both female and male participants equally,  enrolling male participants is significantly more challenging. 
One future direction is developing enrollment strategies to achieve gender balance and analyze the gender balanced datasets to validate our results.    

Our dataset is also imbalanced in the amount of data coming from the iOS and Android platforms. The Android dataset is much smaller than the iOS dataset. As a result, we observed better prediction results for iOS than Android, despite data augmentation to balance the two datasets. Having more balanced data from these two platforms might lead to better results, which needs to be further investigated. 

Despite using the data fusion methodology in~\cite{Yue18:fusion} to handle missing location data, we still observed a significant amount of missing data, which reduces the dataset that we can analyze. More reliable data collection strategies can reduce the amount of missing data. 
Along similar lines, more effective data imputation techniques, particularly those that can handle a large amount of missing data, can be helpful future directions. Last, our data is from a diverse demographic group. Future studies can focus on less diverse groups and potentially develop more effective customized machine learning models. 

%% file: conclusion-SS.tex
\section{Conclusions} \label{sec:conclusions} 
In this paper, we have explored using domain adaptation techniques to align location features collected on iOS and Android platforms to form a larger dataset. Using the larger dataset, we have trained machine learning  models to predict the status of depression treatment. Our results show that this approach leads to accurate prediction for both the iOS and Android datasets. In addition, we show that using location sensory data combined  with baseline self-reported questionnaire score  can lead to $F_1$ score up to 0.67, comparable to the $F_1$ score obtained using periodic self-reported scores.  

%% file: main.bbl
\begin{thebibliography}{10}

\bibitem{Stone02:Capturing}
S.~S. Arthur A~Stone.
\newblock Capturing momentary, self-report data: a proposal for reporting guidelines.
\newblock {\em Annals of behavioral medicine : a publication of the Society of Behavioral Medicine}, 2002.

\bibitem{Shai10:domain-adapt}
S.~Ben-David, J.~Blitzer, K.~Crammer, A.~Kulesza, F.~Pereira, and J.~Wortman~Vaughan.
\newblock A theory of learning from different domains.
\newblock {\em Machine Learning}, 79(1–2), 2010.

\bibitem{Ben-Zeev15L:psychiatric}
D.~Ben-Zeev, E.~A. Scherer, R.~Wang, H.~Xie, and A.~T. Campbell.
\newblock Next-generation psychiatric assessment: Using smartphone sensors to monitor behavior and mental health.
\newblock {\em Psychiatric Rehabilitation Journal}, 38(3):218--226, 2015.

\bibitem{Zeev10:Retrospective}
D.~Ben-Zeev and M.~A. Young.
\newblock Accuracy of hospitalized depressed patients' and healthy controls' retrospective symptom reports: an experience sampling study.
\newblock {\em The Journal of nervous and mental disease}, 198(4):280--285, 2010.

\bibitem{boser1992training}
B.~E. Boser, I.~M. Guyon, and V.~N. Vapnik.
\newblock A training algorithm for optimal margin classifiers.
\newblock In {\em Proceedings of the fifth annual workshop on Computational learning theory}, pages 144--152, 1992.

\bibitem{cacioppo2010perceived}
J.~T. Cacioppo, L.~C. Hawkley, and R.~A. Thisted.
\newblock Perceived social isolation makes me sad: 5-year cross-lagged analyses of loneliness and depressive symptomatology in the {Chicago} health, aging, and social relations study.
\newblock {\em Psychology and aging}, 25(2):453, 2010.

\bibitem{Canzian:2015:TDU:2750858.2805845}
L.~Canzian and M.~Musolesi.
\newblock Trajectories of depression: Unobtrusive monitoring of depressive states by means of smartphone mobility traces analysis.
\newblock In {\em Proc. of ACM UbiComp}, pages 1293--1304, 2015.

\bibitem{CC01a:libSVM}
C.-C. Chang and C.-J. Lin.
\newblock {LIBSVM}: A library for support vector machines.
\newblock {\em ACM Transactions on Intelligent Systems and Technology}, 2:27:1--27:27, 2011.
\newblock Software available at \url{http://www.csie.ntu.edu.tw/~cjlin/libsvm}.

\bibitem{Chekroud21:promise}
A.~Chekroud, J.~Bondar, J.~Delgadillo, and et~al.
\newblock The promise of machine learning in predicting treatment outcomes in psychiatry.
\newblock {\em World Psychiatry}, 20, 2021.

\bibitem{chen2016xgboost}
T.~Chen and C.~Guestrin.
\newblock {XGBoost:} a scalable tree boosting system.
\newblock In {\em Proceedings of ACM SIGKDD International Conference on Knowledge Discovery and Data Mining}, pages 785--794, 2016.

\bibitem{Chikersal21:depression}
P.~Chikersal, A.~Doryab, M.~Tumminia, D.~K. Villalba, J.~M. Dutcher, X.~Liu, S.~Cohen, K.~G. Creswell, J.~Mankoff, J.~D. Creswell, M.~Goel, and A.~K. Dey.
\newblock {Detecting depression and predicting its onset using longitudinal symptoms captured by passive sensing: A machine learning approach with robust feature selection}.
\newblock {\em ACM Transactions on Computer-Human Interaction}, 28(1), February 2021.

\bibitem{Chow17:mobilesensing}
I.~P. Chow, K.~Fua, Y.~Huang, W.~Bonelli, H.~Xiong, E.~L. Barnes, and A.~B. Teachman.
\newblock Using mobile sensing to test clinical models of depression, social anxiety, state affect, and social isolation among college students.
\newblock {\em J Med Internet Res}, 19(3), Mar 2017.

\bibitem{Cohen18:treatment}
Z.~D. Cohen and R.~J. DeRubeis.
\newblock Treatment selection in depression.
\newblock {\em Annu. Rev. Clin. Psychol}, 14(15), 2018.

\bibitem{cortes1995support}
C.~Cortes and V.~Vapnik.
\newblock Support-vector networks.
\newblock {\em Machine learning}, 20(3):273--297, 1995.

\bibitem{Cuijpers02:Excess}
P.~Cuijpers and F.~Smit.
\newblock Excess mortality in depression: a meta-analysis of community studies.
\newblock {\em J Affect Disord}, 72(3):227--236, December 2002.

\bibitem{Dai22:treatment}
R.~Dai, R.~Kannampallil, J.~Zhang, N.~Lv, J.~Ma, and C.~Lu.
\newblock Multi-task learning for randomized controlled trials: A case study on predicting depression with wearable data.
\newblock {\em Proceedings of the ACM on Interactive, Mobile, Wearable and Ubiquitous Technologies (IMWUT)}, 6(2), 2022.

\bibitem{Demasi16:change}
O.~Demasi, A.~Aguilera, and B.~Recht.
\newblock Detecting change in depressive symptoms from daily wellbeing questions, personality, and activity.
\newblock In {\em IEEE Wireless Health}, 2016.

\bibitem{Eid2019:Sexdifferences}
R.~Eid, A.~Gobinath, and L.~Galea.
\newblock Sex differences in depression: Insights from clinical and preclinical studies.
\newblock {\em Prog Neurobiol.}, May 2019.

\bibitem{ester1996density}
M.~Ester, H.-P. Kriegel, J.~Sander, and X.~Xu.
\newblock A density-based algorithm for discovering clusters in large spatial databases with noise.
\newblock In {\em ACM KDD}, volume~96, pages 226--231, 1996.

\bibitem{Farhan16:Multi-view}
A.~A. Farhan, J.~Lu, J.~Bi, A.~Russell, B.~Wang, and A.~Bamis.
\newblock Multi-view bi-clustering to identify smartphone sensing features indicative of depression.
\newblock In {\em Proc. IEEE CHASE}, June 2016.

\bibitem{Farhan16:depression}
A.~A. Farhan, C.~Yue, R.~Morillo, S.~Ware, J.~Lu, J.~Bi, J.~Kamath, A.~Russell, A.~Bamis, and B.~Wang.
\newblock Behavior vs. introspection: Refining prediction of clinical depression via smartphone sensing data.
\newblock In {\em Proc. of Wireless Health}, 2016.

\bibitem{Fortney17:Measurement-Based}
J.~C. Fortney, J.~Unutzer, G.~Wrenn, J.~M. Pyne, G.~R. Smith, M.~Schoenbaum, and H.~T. Harbin.
\newblock A tipping point for measurement-based care.
\newblock {\em Psychiatric Services}, 68(2), February 2017.

\bibitem{frost2013supporting}
M.~Frost, A.~Doryab, M.~Faurholt-Jepsen, L.~V. Kessing, and J.~E. Bardram.
\newblock Supporting disease insight through data analysis: refinements of the monarca self-assessment system.
\newblock In {\em Proceedings of the 2013 ACM international joint conference on Pervasive and ubiquitous computing}, pages 133--142. ACM, 2013.

\bibitem{gruenerbl2014using}
A.~Gruenerbl, V.~Osmani, G.~Bahle, J.~C. Carrasco, S.~Oehler, O.~Mayora, C.~Haring, and P.~Lukowicz.
\newblock Using smart phone mobility traces for the diagnosis of depressive and manic episodes in bipolar patients.
\newblock In {\em Proceedings of the 5th Augmented Human International Conference}, page~38. ACM, 2014.

\bibitem{grunerbl2012towards}
A.~Gr{\"u}nerbl, P.~Oleksy, G.~Bahle, C.~Haring, J.~Weppner, and P.~Lukowicz.
\newblock Towards smart phone based monitoring of bipolar disorder.
\newblock In {\em Proceedings of the Second ACM Workshop on Mobile Systems, Applications, and Services for HealthCare}, page~3. ACM, 2012.

\bibitem{Guy76:CGI}
W.~Guy, editor.
\newblock {\em {ECDEU} Assessment Manual for Psychopharmacology}.
\newblock Rockville, MD: US Department of Heath, Education, and Welfare Public Health Service Alcohol, Drug Abuse, and Mental Health Administration, 1976.

\bibitem{guyon2002gene}
I.~Guyon, J.~Weston, S.~Barnhill, and V.~Vapnik.
\newblock Gene selection for cancer classification using support vector machines.
\newblock {\em Machine learning}, 46(1-3):389--422, 2002.

\bibitem{katevas2016sensingkit}
K.~Katevas, H.~Haddadi, and L.~Tokarchuk.
\newblock Sensingkit: Evaluating the sensor power consumption in ios devices.
\newblock 2016.

\bibitem{katon2002impact}
W.~Katon and P.~Ciechanowski.
\newblock Impact of major depression on chronic medical illness.
\newblock {\em Journal of Psychosomatic Research}, 53(4):859--863, 2002.

\bibitem{Kemp08:improving}
A.~Kemp, E.~Gordon, A.~Rush, and L.~Williams.
\newblock Improving the prediction of treatment response in depression: Integration of clinical, cognitive, psychophysiological, neuroimaging, and genetic measures.
\newblock {\em CNS Spectr.}, 13(12), 2008.

\bibitem{lane2014bewell}
N.~D. Lane, M.~Lin, M.~Mohammod, X.~Yang, H.~Lu, G.~Cardone, S.~Ali, A.~Doryab, E.~Berke, A.~T. Campbell, et~al.
\newblock {BeWell}: Sensing sleep, physical activities and social interactions to promote wellbeing.
\newblock {\em Mobile Networks and Applications}, 19(3):345--359, 2014.

\bibitem{Lathia:2013:OSS:2494091.2497345}
N.~Lathia, K.~Rachuri, C.~Mascolo, and G.~Roussos.
\newblock Open source smartphone libraries for computational social science.
\newblock In {\em Proc. of ACM UbiComp}, UbiComp '13 Adjunct, pages 911--920, 2013.

\bibitem{Lee18:predict}
Y.~Lee, R.~Ragguett, R.~Mansur, and et~al.
\newblock Applications of machine learning algorithms to predict therapeutic outcomes in depression: a meta-analysis and systematic review.
\newblock {\em J Affect Disord}, 2018.

\bibitem{Lu2018:MTL}
J.~Lu, C.~Shang, C.~Yue, R.~Morillo, S.~Ware, J.~Kamath, A.~Bamis, A.~Russell, B.~Wang, and J.~Bi.
\newblock Joint modeling of heterogeneous sensing data for depression assessment via multi-task learning.
\newblock {\em Proceedings of the ACM on Interactive, Mobile, Wearable and Ubiquitous Technologies (IMWUT)}, 2(1), 2018.

\bibitem{martinsen1990benefits}
E.~W. Martinsen.
\newblock Benefits of exercise for the treatment of depression.
\newblock {\em Sports Medicine}, 9(6):380--389, 1990.

\bibitem{Mehrotra16:Towards}
A.~Mehrotra, R.~Hendley, and M.~Musolesi.
\newblock Towards multi-modal anticipatory monitoring of depressive states through the analysis of human-smartphone interaction.
\newblock In {\em Proc. of UbiComp}, 2016.

\bibitem{meyerhoff2021evaluation}
J.~Meyerhoff, T.~Liu, K.~P. Kording, L.~H. Ungar, S.~M. Kaiser, C.~J. Karr, D.~C. Mohr, et~al.
\newblock Evaluation of changes in depression, anxiety, and social anxiety using smartphone sensor features: longitudinal cohort study.
\newblock {\em Journal of medical Internet research}, 23(9):e22844, 2021.

\bibitem{Mohr17:sensing}
D.~C. Mohr, M.~Zhang, and S.~M. Schueller.
\newblock Personal sensing: understanding mental health using ubiquitous sensors and machine learning.
\newblock {\em Annu Rev Clin Psychol}, 2017.

\bibitem{Morris12:Measurement-Based}
D.~W. Morris, M.~Toups, and M.~H. Trivedi.
\newblock Measurement-based care in the treatment of clinical depression.
\newblock {\em FOCUS: The Journal of Lifelong Learning in Psychiatry}, 2012.

\bibitem{Palmius16Detecting}
N.~Palmius, A.~Tsanas, K.~E.~A. Saunders, A.~C. Bilderbeck, J.~R. Geddes, G.~M. Goodwin, and M.~D. Vos.
\newblock Detecting bipolar depression from geographic location data.
\newblock {\em IEEE Transactions on Biomedical Engineering}, 64(8):1761--1771, 2017.

\bibitem{Pan10:transfer-learning}
S.~J. Pan and Q.~Yang.
\newblock A survey on transfer learning.
\newblock {\em IEEE Transactions on Knowledge and Data Engineering}, 2010.

\bibitem{doi:10.1056/NEJMp2008017}
B.~Pfefferbaum and C.~S. North.
\newblock Mental health and the covid-19 pandemic.
\newblock {\em New England Journal of Medicine}, 383(6):510--512, 2020.
\newblock PMID: 32283003.

\bibitem{rakotomamonjy2003variable}
A.~Rakotomamonjy.
\newblock Variable selection using {SVM}-based criteria.
\newblock {\em Journal of machine learning research}, 3(Mar):1357--1370, 2003.

\bibitem{Redko19:domain-adapt}
I.~Redko, E.~Morvant, A.~Habrard, M.~Sebban, and Y.~Bennani.
\newblock {\em Advances in Domain Adaptation Theory}.
\newblock ISTE Press - Elsevier, 2019.

\bibitem{Darius18:Correlations}
D.~A. Rohani, M.~Faurholt-Jepsen, L.~V. Kessing, and J.~E. Bardram.
\newblock Correlations between objective behavioral features collected from mobile and wearable devices and depressive mood symptoms in patients with affective disorders: Systematic review.
\newblock {\em JMIR Mhealth Uhealth}, 2018.

\bibitem{Rost22:treatment}
N.~Rost, E.~B. Binder, and T.~M. Brückl.
\newblock Predicting treatment outcome in depression: an introduction into current concepts and challenges.
\newblock {\em European Archives of Psychiatry and Clinical Neuroscience}, 2022.

\bibitem{rush200316}
A.~J. Rush, M.~H. Trivedi, H.~M. Ibrahim, T.~J. Carmody, B.~Arnow, D.~N. Klein, J.~C. Markowitz, P.~T. Ninan, S.~Kornstein, R.~Manber, et~al.
\newblock The 16-item quick inventory of depressive symptomatology {(QIDS)}, clinician rating {(QIDS-C)}, and self-report {(QIDS-SR)}: a psychometric evaluation in patients with chronic major depression.
\newblock {\em Biological psychiatry}, 54(5):573--583, 2003.

\bibitem{Saeb16:student-life-data}
S.~Saeb, E.~G. Lattie, S.~M. Schueller, K.~P. Kording, and D.~C. Mohr.
\newblock The relationship between mobile phone location sensor data and depressive symptom severity.
\newblock {\em PeerJ}, 4(e2537), 2016.

\bibitem{saeb2015mobile}
S.~Saeb, M.~Zhang, C.~J. Karr, S.~M. Schueller, M.~E. Corden, K.~P. Kording, and D.~C. Mohr.
\newblock Mobile phone sensor correlates of depressive symptom severity in daily-life behavior: An exploratory study.
\newblock {\em Journal of Medical Internet Research}, 17(7), 2015.

\bibitem{sanders2000relationship}
C.~E. Sanders, T.~M. Field, D.~Miguel, and M.~Kaplan.
\newblock The relationship of {Internet} use to depression and social isolation among adolescents.
\newblock {\em Adolescence}, 35(138):237, 2000.

\bibitem{Shende2023:PredictingSymptom}
C.~Shende, S.~Sahoo, S.~Sam, P.~Patel, R.~Morillo, X.~Wang, S.~Ware, J.~Bi, J.~Kamath, A.~Russell, and D.~Song.
\newblock Predicting symptom improvement during depression treatment using sleep sensory data.
\newblock {\em Proceedings of the ACM on Interactive, Mobile, Wearable and Ubiquitous Technologies}, 7:1--21, 09 2023.

\bibitem{Simon10:personalized}
G.~E. Simon and R.~H. Perlis.
\newblock Personalized medicine for depression: Can we match patients with treatments.
\newblock {\em Am J Psychiatry}, 167(12), December 2010.

\bibitem{Sahoo2024:DailyMood}
M.~Z.~H. Soumyashree~Sahoo, Chinmaey~Shende.
\newblock Using mobile daily mood and anxiety self-ratings to predict depression symptom improvement.
\newblock {\em 2024 IEEE/ACM International Conference on Connected Health: Applications, Systems and Engineering Technologies (CHASE)}, 2024.

\bibitem{Spitzer99:PHQ-9}
R.~Spitzer, K.~Kroenke, and J.~Williams.
\newblock Validation and utility of a self-report version of {PRIME-MD}: the {PHQ} primary care study. primary care evaluation of mental disorders. patient health questionnaire.
\newblock {\em JAMA}, 282(18):1737--1744, 1999.

\bibitem{Suhara16:DeepMood}
Y.~Suhara, Y.~Xu, and A.~Pentland.
\newblock Deepmood: Forecasting depressed mood based on self-reported histories via recurrent neural networks.
\newblock {\em Proc. of WWW}, 2017.

\bibitem{Sun15:CORAL}
B.~Sun, J.~Feng, and K.~Saenko.
\newblock Return of frustratingly easy domain adaptation.
\newblock {\em CoRR}, abs/1511.05547, 2015.

\bibitem{Sun16a:DeepCORAL}
B.~Sun and K.~Saenko.
\newblock Deep {CORAL:} correlation alignment for deep domain adaptation.
\newblock {\em CoRR}, abs/1607.01719, 2016.

\bibitem{07648d6cc15940829f39edccd66719ab}
{Van Marwijk}, Harm, Mitchell, {Alex J}, A.~Vaze, and S.~Rao.
\newblock Clinical diagnosis of depression in primary care: a meta-analysis.
\newblock {\em The Lancet}, 374:609--619, 2009.

\bibitem{Wang16:schizophrenia}
R.~Wang, M.~S.~H. Aung, S.~Abdullah, R.~Brian, A.~T. Campbell, T.~Choudhuryy, M.~Hauserz, J.~Kanez, M.~Merrilly, E.~A. Scherer, V.~W.~S. Tsengy, and D.~Ben-Zeev.
\newblock Crosscheck: Toward passive sensing and detection of mental health changes in people with schizophrenia.
\newblock In {\em Proc. of UbiComp}, 2016.

\bibitem{wang2014studentlife}
R.~Wang, F.~Chen, Z.~Chen, T.~Li, G.~Harari, S.~Tignor, X.~Zhou, D.~Ben-Zeev, and A.~T. Campbell.
\newblock Studentlife: assessing mental health, academic performance and behavioral trends of college students using smartphones.
\newblock In {\em Proceedings of the 2014 ACM International Joint Conference on Pervasive and Ubiquitous Computing}, pages 3--14. ACM, 2014.

\bibitem{Wang18:dynamics}
R.~Wang, W.~Wang, A.~daSilva, J.~F. Huckins, W.~M. Kelley, T.~F. Heatherton, and A.~T. Campbell.
\newblock Tracking depression dynamics in college students using mobile phone and wearable sensing.
\newblock {\em Proceedings of the ACM on Interactive, Mobile, Wearable and Ubiquitous Technologies}, 2(1), 2018.

\bibitem{Myrna2014:Treatment}
M.~M. Weissman.
\newblock Treatment of depression: Men and women are different?
\newblock {\em The AMerican Journal of Psychiatry.}, 2014.

\bibitem{Xiong2016SensusAC}
H.~Xiong, Y.~Huang, L.~E. Barnes, and M.~S. Gerber.
\newblock Sensus: a cross-platform, general-purpose system for mobile crowdsensing in human-subject studies.
\newblock {\em Proceedings of the 2016 ACM International Joint Conference on Pervasive and Ubiquitous Computing}, 2016.

\bibitem{yan2015feature}
K.~Yan and D.~Zhang.
\newblock Feature selection and analysis on correlated gas sensor data with recursive feature elimination.
\newblock {\em Sensors and Actuators B: Chemical}, 212:353--363, 2015.

\bibitem{Yue18:fusion}
C.~Yue, S.~Ware, R.~Morillo, J.~Lu, C.~Shang, J.~Bi, A.~Russell, A.~Bamis, and B.~Wang.
\newblock Fusing location data for depression prediction.
\newblock {\em IEEE Transactions on Big Data}, 2018.

\bibitem{Yuuki2020:Aware}
Y.~E. Yuuki~Nishiyama, Denzil~Ferreira.
\newblock {IOS} crowd–sensing won’t hurt a bit!: {AWARE} framework and sustainable study guideline for {iOS} platform.
\newblock {\em Distributed, Ambient and Pervasive Interactions: 8th International Conference}, 2020.

\bibitem{Zhang21:Relationship}
Y.~Zhang, A.~A. Folarin, S.~Sun, N.~Cummins, R.~Bendayan, Y.~Ranjan, Z.~Rashid, P.~Conde, C.~Stewart, P.~Laiou, F.~Matcham, K.~M. White, F.~Lamers, S.~Siddi, S.~Simblett, I.~Myin-Germeys, A.~Rintala, T.~Wykes, J.~M. Haro, B.~W. Penninx, V.~A. Narayan, M.~Hotopf, and R.~J. Dobson.
\newblock Relationship between major depression symptom severity and sleep collected using a wristband wearable device: Multicenter longitudinal observational study.
\newblock {\em JMIR Mhealth Uhealth}, 9(4), Apr 2021.

\bibitem{Zhou15:tackling}
D.~Zhou, J.~Luo, V.~M.~B. Silenzio, Y.~Zhou, J.~Hu, G.~Currier, and H.~A. Kautz.
\newblock Tackling mental health by integrating unobtrusive multimodal sensing.
\newblock In {\em Proc. of AAAI}, 2015.

\bibitem{Zou23:Sequence}
B.~Zou, X.~Zhang, L.~Xiao, R.~Bai, X.~Li, H.~Liang, H.~Ma, and G.~Wang.
\newblock Sequence modeling of passive sensing data for treatment response prediction in major depressive disorder.
\newblock {\em IEEE Transactions on Neural Sciences and Rehabilitation Engineering}, 31, 2023.

\end{thebibliography}
